\documentclass[10pt,twocolumn,letterpaper]{article}

\usepackage[pagenumbers]{wacv} 
\definecolor{wacvblue}{rgb}{0.21,0.49,0.74}
\usepackage[pagebackref,breaklinks,colorlinks,allcolors=wacvblue]{hyperref}
\usepackage{multirow,multicol}
\usepackage{pifont}
\usepackage{array}

\newcolumntype{C}[1]{>{\centering\arraybackslash}p{#1}}

\title{Beyond Clear Skies: Synthetic Seasonal and Weather Variations for Real-World Drone Detection}

\author{
    Tamara R. Lenhard$^{1,2}$ 
    \and
    Andreas Weinmann$^{2}$
    \and
    Tobias Koch$^{1}$\\\vspace{-0.4cm}\and
    {\normalsize$^{1}$Institute for the Protection of Terrestrial Infrastructures,
    German Aerospace Center (DLR), Sankt Augustin, Germany}\\
    {\normalsize$^{2}$ACIDA Lab, Technical University of Applied Sciences Würzburg-Schweinfurt, Schweinfurt, Germany}\\\vspace{-0.4cm}\and
    {\tt\small \{tamara.lenhard, tobias.koch\}@dlr.de, andreas.weinmann@thws.de} \vspace{-0.4cm}
}

\begin{document}
\maketitle

\begin{abstract}
Reliable drone detection under real-world deployment conditions requires training data that spans the full operational design domain, including adverse weather and seasonal appearance variation. However, acquiring and annotating such data at scale remains highly resource-intensive, as adverse-weather conditions are inherently difficult to control, reproduce, and sample systematically. Existing datasets therefore typically provide only limited coverage of such conditions. Conversely, synthetic data offers a scalable alternative: environmental variation becomes controllable, while modern game-engine-based pipelines provide realistic rendering and automatic annotations. Leveraging this potential, we introduce SynDroneVision-Weather (SDV-W), a systematic extension of SynDroneVision (SDV) targeting adverse-weather and seasonal domain shifts in urban drone detection. SDV-W comprises 55,187 annotated high-resolution images of urban environments, rendered across three seasonal configurations and diverse weather conditions, including rain, snow, and fog at multiple severity levels. By preserving SDV's scene and trajectory configuration, SDV-W enables controlled, condition-specific assessment of detector degradation through paired clear/adverse frames. Evaluations across representative YOLO models and real-world datasets demonstrate that SDV-W improves detection performance by reducing both missed detections and false alarms, and is most effective as a complement to general-purpose synthetic drone-detection data. SDV-W will be publicly released upon paper acceptance.
\end{abstract}

\section{Introduction}
\label{sec:intro}
Unmanned aerial vehicles (UAVs) -- more commonly known as drones -- have evolved from specialized military systems into versatile aerial platforms, supporting applications in logistics, agriculture, infrastructure inspection, surveillance, and recreation~\cite{Mohsan:2023}. However, this increased accessibility also raises security and privacy concerns, as non-cooperative UAV operations can compromise airspace integrity, operational safety, and critical-infrastructure protection~\cite{Dafrallah:2024}. Consequently, reliable drone detection is essential for low-altitude airspace surveillance, enabling timely localization and risk mitigation. Camera-based sensing provides a broadly deployable modality for drone detection~\cite{Dong:2025}, particularly in urban environments where active sensing technologies are frequently affected by electromagnetic interference, environmental clutter, and site-specific deployment constraints~\cite{Seidaliyeva:2024}. Coupled with modern \textit{deep learning (DL)} techniques, camera-based sensing offers a scalable and cost-effective strategy for drone detection~\cite{Elsayed:2021}.

The reliability of DL-based detectors is inherently tied to the diversity and representativeness of their training data~\cite{Singh:2024,Yu:2023}. In practice, this requires comprehensive coverage of the detector's \textit{operational design domain (ODD)} -- the range of visual, environmental, and operational conditions under which reliable detection must be maintained. Conditions outside this domain can induce substantial performance degradation~\cite{Torens:2025}. Accordingly, real-world drone detection depends on training data that captures both conventional visual variability -- viewpoint, target scale, background structure, and illumination -- and operationally relevant distribution shifts induced by seasonality and adverse weather, which jointly affect scene appearance and target-background contrast. However, acquiring and annotating real-world data across the full ODD remains costly and difficult to control~\cite{Jiang:2023,Svanstroem:2021}, particularly under adverse weather conditions. Consequently, existing datasets often lack sufficient coverage of these scenarios, constraining their ability to support robust generalization under deployment-relevant distribution shifts.

\begin{table*}[t!]
\centering\footnotesize

\caption{\label{tab:dronedatasets_overview}Overview of image-based drone-detection datasets in terms of adverse-weather representation and seasonal coverage. Image origin denotes whether samples are directly acquired from real scenes~(R), generated by image-space augmentation of real images~(A), or rendered synthetically~(S). The symbols \ding{51} and \ding{55} indicate whether diverse drone models or adverse-weather phenomena are represented. Weather coverage refers to visually observable atmospheric effects, such as falling rain or snow, rather than residual scene conditions (\eg, snow covered ground). Scene context characterizes the visual setting of adverse-weather phenomena. Season denotes dataset-wide seasonal coverage, independent of weather conditions (SP: spring, SU: summer, AU: autumn, WI: winter). ``--'' indicates information not reported or inferable from the available data.\vspace{-0.2cm}}
  \begin{tabular}{@{}|l|c|c|c|c|c|c|c|c|c|c|@{}}
    \hline
    Dataset & Img & No. & Max. Img. & Diff. &  \multicolumn{4}{c|}{Adverse Weather} & Season \\
    & Origin & Imgs & Resolution & Drones & Rain & Snow & Fog & Scene Context &\\\hline
    
    USC~\cite{Chen:2017,Wang:2018}  & R & 27,000 & 1920$\times$1080 & \ding{55} & -- & -- & -- & -- & -- \\\hline
    
    Drone Dataset~\cite{Aksoy:2019} & R &  4,012 & 3840$\times$2160 & \ding{55} & \ding{55} & \ding{55} & \ding{55} & -- &  SP/SU, AU\\\hline

    UAV-Eagle~\cite{Barisic:2021} & R &  510 & 1920$\times$1080 & \ding{55} & \ding{55} & \ding{55} & \ding{55} & -- &  SP/SU\\\hline
    
    UAVData~\cite{Zeng:2021} & R & 13,803 & 1280$\times$720 & \ding{51} & \ding{55} & \ding{55} & \ding{55} & -- &  SP/SU\\\hline
    
    Halmstadt Data~\cite{Svanstroem:2021} & R & 203,328$^{\text{\ding{115}}}$ & 640$\times$512 & \ding{51} & \ding{55} & \ding{55} & \ding{55} & -- & -- \\\hline
    
    DUT Anti-UAV~\cite{Zhao:2022} & R & 10,000 & 5616$\times$3744 & \ding{51} & \ding{55} & \ding{55} & \ding{55} & -- & SP/SU, WI\\ \hline
    
    VisioDECT~\cite{Ajakwe:2022} & R & 20,924 & 852$\times$480 & \ding{51} & \ding{55} & \ding{55} & \ding{55} & -- & -- \\\hline

    Drone-vs-Bird Detection~\cite{Coluccia:2024} & R & 85,904 & 3840$\times$2160 & \ding{51} & \ding{55} & \ding{55} & \ding{55} & -- & --\\ \hline
    
    LRDDv1$^{\text{\ding{72}}}$~\cite{Rouhi:2024} & R & 22,342 & 1920$\times$1080 & \ding{51} & \ding{55} & \ding{55} & \ding{55} & -- & SU, WI\\\hline
    
    LRDDv2$^{\text{\ding{72}}}$~\cite{Rouhi:2025} & R & 16,396 & 1920$\times$1080 & \ding{51} & \ding{51}$^{\text{\ding{68}}}$ & \ding{55} & \ding{55} & urban & SU, WI\\\hline
    
    LRDDv3$^{\text{\ding{72}}}$~\cite{Peterson:2026} & R & 102,532 & 3840$\times$2160 & \ding{55} & \ding{51}$^{\text{\ding{68}}}$ & \ding{51}$^{\text{\ding{68}}}$ & \ding{55} & urban & SU, WI\\\noalign{\hrule height 0.8pt}
    
    Rainy Test Set (RTS)~\cite{Munir:2023,Munir:2024} & A & 7,584 & 1020$\times$720 & \ding{51} & \ding{51} & \ding{55} & \ding{55} & mixed & SP/SU, WI\\\hline
    
    S-UAV-T~\cite{Barisic:2022} & S & \ \ 52,500 & 608$\times$608 & \ding{51} & \ding{55} & \ding{55} & \ding{55} & -- & --\\ \hline
    
    DrIFT~\cite{Dadboud:2025} & R$+$S & 47,991 & 1920$\times$1080 & \ding{51} & \ding{51} & \ding{51} & \ding{51} & sky only & SU, AU, WI \\\hline
    
    \multirow{2}{*}{RV-DroneEye$^{\text{\ding{61}}}$~\cite{Adiputra:2026}} & \multirow{2}{*}{S} & \multirow{2}{*}{--} & \multirow{2}{*}{--} & \multirow{2}{*}{\ding{51}} & \multirow{2}{*}{\ding{55}} & \multirow{2}{*}{\ding{55}} & \multirow{2}{*}{\ding{51}} & urban, forest, & \multirow{2}{*}{--} \\
    & & & & & & & & lake &  \\\noalign{\hrule height 0.8pt}

    SynDroneVision~\cite{Lenhard:2025} & S & 140,038 & 2560$\times$1489 & \ding{51} & \ding{55} & \ding{55} & \ding{55} & -- & SP/SU\\\hline
    
    \textbf{SynDroneVision-Weather (ours)} & S & 55,187 & 2560$\times$1477 & \ding{51} & \ding{51} & \ding{51} & \ding{51}$^{\text{\ding{117}}}$ & urban, sky& SP/SU, AU, WI \\\hline
    
    \multicolumn{10}{l}{
    {\footnotesize\ding{115}\, RGB$+$IR}
    \hspace{0.3cm}
    {\footnotesize\ding{72}\, available upon request}
    \hspace{0.3cm}
    {\footnotesize\ding{61}\, not publicly available; included for completeness}
    \hspace{0.3cm}
    {\footnotesize\ding{117}\, compound rain$+$fog}}\\
    \multicolumn{10}{l}{
    {\footnotesize\ding{68}\, primarily represented as scene state (\eg, overcast sky, snow-covered ground), not visible precipitation}
    }\vspace{-0.2cm}
    
  \end{tabular}
\end{table*}

Synthetic data generation offers a scalable means of improving ODD coverage. In particular, game-engine-based pipelines expose deployment-relevant factors as controllable parameters, enabling systematic variation of visual conditions while automatically providing pixel-precise annotations. \textit{SynDroneVision (SDV)}~\cite{Lenhard:2025} demonstrates this potential in drone detection, improving performance and cross-domain robustness (especially when complemented by small amounts of real-world data). Nevertheless, its environmental coverage remains limited to variations in illumination, cloud structure, sky appearance, and clear or overcast conditions, excluding adverse weather and seasonal shifts. Existing weather-aware datasets address this gap only partially, either relying on physically inconsistent augmentation~\cite{Munir:2023}, covering a rather narrow range of conditions~\cite{Adiputra:2026}, or being designed primarily for evaluation rather than training~\cite{Dadboud:2025}. Motivated by these limitations, we make the following \textbf{contributions}:
\begin{itemize}
    \item We introduce \textit{SynDroneVision-Weather (SDV-W)}, an extension of SDV~\cite{Lenhard:2025} that re-renders curated SDV sequences across controlled seasonal appearance states and adverse-weather conditions while preserving scene geometry, camera placement, and drone trajectories.\vspace{0.1cm}
    \item We provide a controlled analysis of adverse-weather robustness, quantifying condition-specific degradation across representative detector architectures without explicit weather exposure.\vspace{0.1cm}
    \item We quantify the contribution of SDV-W to cross-domain generalization across diverse drone detectors and real-world datasets covering seasonal appearance variation.\vspace{0.1cm}
    \item We disentangle the effects of environmental diversity and training-set scale, and characterize detector behavior across object size.\vspace{0.1cm}
\end{itemize}

The remainder is organized as follows: Sec.~\ref{sec:related_work} reviews related work on image-based drone detection and publicly available datasets. Sec.~\ref{sec:SDVW} presents SDV-W, including its generation process, key properties, and composition. The experimental setup is outlined in Sec.~\ref{sec:exp_setup}, followed by the results in Sec.~\ref{sec:results}. Conclusions are provided in Sec.~\ref{sec:conclusion}. Further details are available in the supplementary material (Supp.).

\section{Related Work}
\label{sec:related_work}
This section outlines recent advances in image-based drone detection and examines public datasets in terms of seasonal appearance variation and adverse-weather coverage.

\subsection{Drone Detection}
\label{subsec:drone_detection_models}
Image-based drone detection primarily relies on single-stage DL architectures, which offer a favorable compromise between real-time inference and detection accuracy. Among single-stage architectures, the YOLO family remains the prevailing choice, employed either in standard configurations~\cite{Barisic:2022,Munir:2023,Adiputra:2026,Rouhi:2024,Rouhi:2025} or with task-specific modifications~\cite{Lv:2022,Lenhard:2024,Lenhard:2026,Huang:2024}. Architectural adaptations typically address domain-specific challenges, including small-scale or long-range drone detection~\cite{Rouhi:2024,Rouhi:2025,Huang:2024,Lv:2022,Jiang:2023,Liu:2021}, discrimination from visually similar aerial objects such as birds~\cite{Coluccia:2024,Lv:2022}, and robustness to camouflage effects~\cite{Lenhard:2024,Lenhard:2026}. The latter poses a particular challenge in urban surveillance, where structural and vegetative clutter can reduce target-background separability, lowering drone saliency and degrading the performance of generic YOLO-based detectors~\cite{Lenhard:2024}. A representative approach addressing this challenge is \textit{YOLO-FEDER FusionNet}~\cite{Lenhard:2026}. Transformer-based detectors (\eg, RT-DETR~\cite{Zhao:2024,Lv:2024}) provide an alternative to YOLO-centric detection architectures~\cite{Adiputra:2026,Jamil:2022}, yet remain less common in this domain. Despite these advances, detector evaluation rarely accounts for adverse weather or seasonal appearance shifts.

\subsection{Datasets}
\label{subsec:RW_data}
Training DL-based drone detection models predominantly relies on application-specific real-world data, whose acquisition remains labor-intensive, difficult to scale, and operationally constrained~\cite{Jiang:2023,Svanstroem:2021}. These constraints are amplified under adverse weather conditions, where limited experimental control makes systematic data acquisition difficult. Consequently, existing real-world drone detection datasets capture, at most, seasonal variation under clear-to-overcast conditions -- as in LRDDv1~\cite{Rouhi:2024}, LRDDv2~\cite{Rouhi:2025}, and LRDDv3~\cite{Peterson:2026} -- while largely excluding visible adverse atmospheric effects. Although LRDDv2 and LRDDv3 report rainy and snowy conditions (Tab.~\ref{tab:dronedatasets_overview}), these are primarily reflected in scene characteristics (\eg, overcast skies or snow-covered ground) rather than visible precipitation.

Visible adverse-weather effects are therefore introduced almost exclusively through synthetic mechanisms, including image-space augmentation~\cite{Munir:2023} and simulation-based rendering~\cite{Dadboud:2025,Adiputra:2026} -- at least in the context of drone detection. Image-space augmentation superimposes synthetic weather effects, such as rain streaks, onto real images~\cite{Munir:2023}, but lacks physical coupling to scene geometry and illumination, often yielding salient streak artifacts and chromatic shifts (Fig.~\ref{fig:RW_ExamplesOtherData}, Supp.). Simulation-based rendering provides greater environmental control, yet existing datasets remain limited in scope: DrIFT~\cite{Dadboud:2025} confines adverse-weather variation to synthetic sky-background frames, whereas RV-DroneEye\footnote{Currently not publicly available.}\cite{Adiputra:2026} broadens scene coverage to urban, forest, and lake environments but remains restricted to clear and foggy conditions. Moreover, RV-DroneEye relies on diffusion-based post-processing to compensate for limited rendering fidelity. Nevertheless, synthetic data generation offers a scalable alternative to real-world data acquisition, enabling controlled variation of individual factors while isolating their effects under otherwise fixed scene conditions. Prior work has further demonstrated the effectiveness of large-scale synthetic datasets for drone detection~\cite{Lenhard:2025}. These benefits, however, remain constrained by the simulation-reality gap, with data fidelity and composition ultimately governing real-world transfer. Combining synthetic and real data thus offers a practical means of mitigating this gap~\cite{Dieter:2023}.

\begin{figure}[t!]
\centering
  \includegraphics[width=0.42\textwidth, trim={0.2cm 20.9cm 10cm 0cm}, clip]{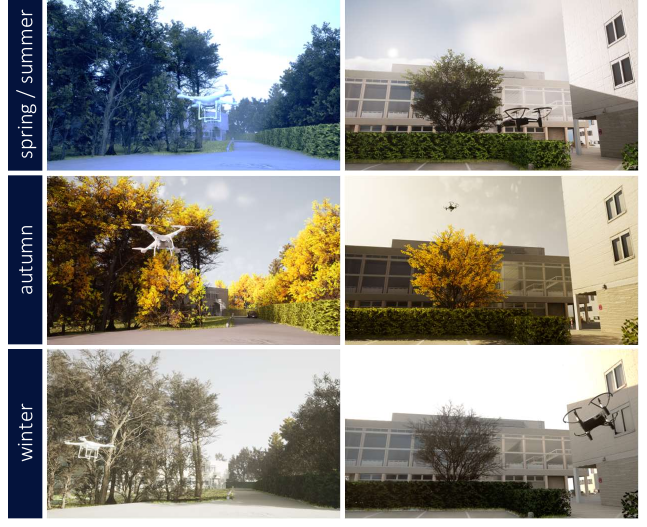}
  \caption{\label{fig:Example_SDV-W_Seasons}Seasonal appearance variation in SDV-W. Identical scenes are re-rendered across seasons, altering vegetation, illumination, and atmospheric appearance while preserving geometry.}
\end{figure}

\section{SynDroneVision-Weather}
\label{sec:SDVW}
This section introduces the proposed SynDroneVision-Weather (SDV-W) dataset, detailing its generation methodology, composition, and key characteristics.

\subsection{Generation Methodology}
SDV-W extends SDV by re-rendering a curated subset of 15 sequences of the original 72-sequence dataset under controlled seasonal and meteorological conditions. Camera placement, drone trajectories, rendering configuration, and pixel-accurate annotations are preserved by reusing the original SDV data generation pipeline~\cite{Lenhard:2025}, built on Colosseum~\cite{Colosseum} and Unreal Engine~5.0 (UE5)~\cite{Unreal}. 

\begin{figure*}[t!]
\centering
  \includegraphics[width=0.96\textwidth, trim={0.1cm 24.3cm 0.1cm 0cm}, clip]{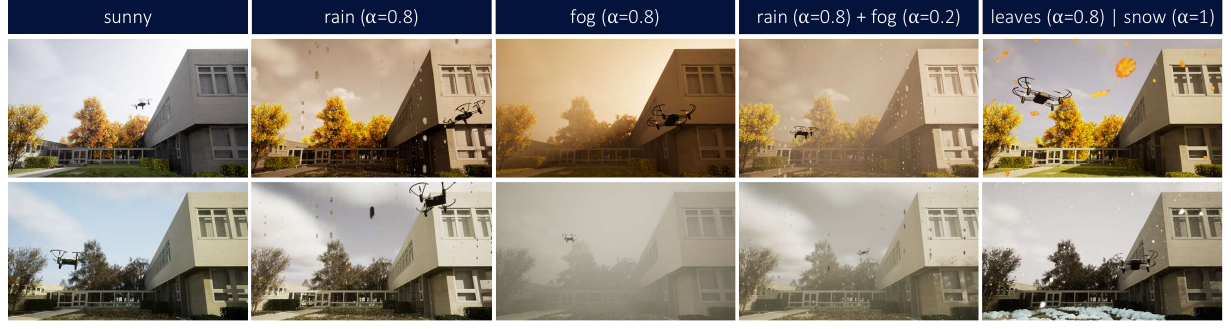}
  \caption{\label{fig:Example_SDV-W_Weather} Weather-condition variation in SDV-W. Representative autumn (top) and winter (bottom) scenes are shown under sunny, rainy, foggy, combined rain$+$fog, and season-specific conditions (leaves/snow). All non-sunny conditions are shown at heavy severity.
}
\end{figure*}

\vspace{-0.3cm}
\paragraph{Environments \& Seasonal Configuration.}
The selected sequences cover three distinct environments: University Site, Venetian City~\cite{Unreal:Venice}, and Urban Downtown~\cite{Unreal:DowntownWest}, each described in detail in~\cite{Lenhard:2025}. While SDV additionally comprises non-urban environments, SDV-W deliberately focuses on urban scenes, reflecting the operational context of urban drone surveillance. Moreover, all three environments contain visually prominent vegetation -- a perceptually dominant indicator of seasonality -- providing a natural basis for inducing controlled seasonal appearance variation (unlike sparsely vegetated settings, such as industrial sites). Each environment is instantiated under three seasonal configurations: spring/summer, autumn, and winter, ex\-pressed primarily through full green, yellow-brown, and defoliated vegetation states, respectively (Fig.~\ref{fig:Example_SDV-W_Seasons}). Spring and summer are treated as a single full-foliage configuration, given their subtle visual difference in urban scenes. This configuration also reflects the default appearance of the underlying environments. The remaining seasonal states are derived via asset-dependent vegetation edits, ranging from native foliage presets and material-level adjustments to seasonally appropriate vegetation replacement (see Supp. for details).

\vspace{-0.3cm}
\paragraph{Weather Conditions.} Weather variation is introduced through the Colosseum weather API~\cite{Colosseum}, which renders precipitation, fog, and particle effects directly within the simulated environment (excluding sensor-side artifacts such as lens droplets). Weather effects are specified by the Colosseum variables \texttt{Rain}, \texttt{Fog}, \texttt{Snow}, and \texttt{MapleLeaf}. Each effect is parameterized by an intensity scalar $\alpha \!\in\! [0,1]$, controlling its strength ($\alpha\!\!=\!\!0$: no effect; $\alpha\!\!=\!\!1$: full in\-tensity). The resulting configurations capture complementary sources of visual variation, combining weather-driven visibility changes with season-specific appearance cues (Fig.~\ref{fig:Example_SDV-W_Weather}). While rain, fog, and their combination are rendered across all seasons, snow and falling foliage are restricted to their respective seasonal contexts. Additional clear-weather scenes (denoted as \emph{sunny}) are generated for autumn and winter to extend the clear-weather coverage beyond the spring/summer baseline provided by SDV. 

Non-sunny conditions are rendered at two severity levels: light ($\alpha \!\in\! [0.2,0.4]$) and heavy ($\alpha \!\in\! [0.8,1.0]$). The compound rain-fog condition constitutes the only exception and is parameterized asymmetrically, with rain set to $\alpha\!\!=\!\!0.8$ and fog to $\alpha\!\!=\!\!0.2$. This asymmetric combination is designed to better approximate realistic rainy-scene conditions. The rain component preserves explicit precipitation cues, while fog reproduces the diffuse illumination, reduced contrast, and desaturation commonly associated with rainfall.

While the remaining weather effects follow the default Colosseum configuration, rain required dedicated particle-material calibration to ensure consistent droplet visibility and visually plausible appearance across varying lighting conditions. Accordingly, the \texttt{P\_Weather\_RainFX} emitter was modified by setting the initial droplet-size distribution of \texttt{RainDropsGPU} to a uniform vector range of $[4,150,4]$-$[15,50,15]$. The \texttt{M\_RainDrop\_Inst} material was calibrated per sequence by adjusting emissive intensity ($0.005$-$1.0$) and metallic reflectance ($0$-$0.4$) to maintain realistic specular response under scene-specific lighting.

\vspace{-0.3cm}
\paragraph{Illumination Settings.} Following the SDV illumination setup~\cite{Lenhard:2025}, scene appearance under each weather-season configuration is defined through the joint calibration of three UE5 components: (i)~the Sun and Sky Actor~\cite{Unreal:SunSkyActor}, controlling solar direction, directional-light intensity, and sky-dependent ambient illumination; (ii)~the Post Process Volume~\cite{Unreal:PostProcessVol}, adjusting global color temperature and tint; and (iii)~the Volumetric Cloud Component~\cite{Unreal:VolumetricCloud}, governing cloud density, spatial structure, and overcast coverage. Note that the Urban Downtown environment~\cite{Unreal:DowntownWest} retains its native Directional Light~\cite{Unreal:DirectionalLight} and Sky Light~\cite{Unreal:SkyLight} configuration instead of relying on the Sun and Sky Actor. Parameter ranges follow each environment's native lighting setup, with full configurations provided in Sec.~\ref{subsec:illumination_params} (Supp.). Adverse weather conditions are rendered relative to the sunny baseline through reduced directional-light intensity, cooler and less saturated color grading, and increased cloud extinction.

\vspace{-0.3cm}
\paragraph{Frame Sampling \& Re-rendering.} Given the calibrated seasonal, weather, and illumination configurations, SDV-W is generated by re-rendering selected frames from the original SDV trajectories. For each of the 15 sequences and each combination of season, weather condition, and severity level, $\sim$200 frames are uniformly sampled along the pre-recorded drone trajectory and re-rendered with the corresponding scene parameters. Preserving the camera intrinsics and trajectory waypoints from SDV ensures a strict geometric correspondence between SDV-W frames and their clean-condition counterparts. Drone pose, viewpoint, and scene structure remain identical across conditions (Fig.~\ref{fig:supp_example_pairing}, Supp.). This one-to-one correspondence isolates seasonal and atmospheric appearance changes from variations in scene content, providing a level of control rarely achievable in real-world recordings. Sampling across the full trajectory further retains intra-sequence diversity in drone scale, viewing perspectives, and object-background relationships. 

\subsection{Dataset Composition}
SDV-W comprises 55,187 annotated single-drone images, distributed across spring/summer (12,594), autumn (20,997), and winter (21,596). A detailed breakdown of image counts by season and weather condition is provided in Tab.~\ref{tab:SDV_Weather_Stats}. Moreover, the dataset is partitioned into training and validation splits of 52,427 and 2,760 images, respectively. 

\begin{figure}[t!]
\centering
  \includegraphics[width=0.4\textwidth, trim={7.2cm 23.4cm 5.2cm 0.2cm}, clip]{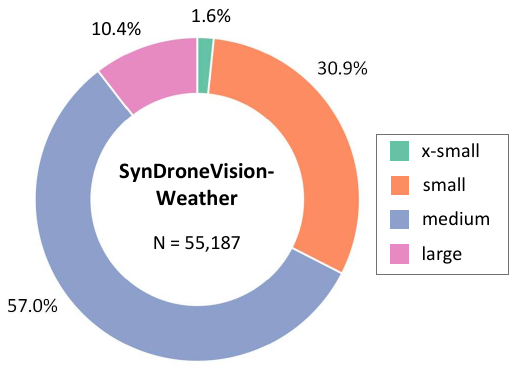}
  \caption{\label{fig:Dist_ObjectSizeCat_SDVW} Distribution of object size categories (Tab.~\ref{tab:obj_size_categories}, Supp.) in SDV-W. Share of x-small, small, medium, and large objects are shown w.r.t. its total number of annotated instances (N).}
\end{figure}

\subsection{Dataset Characteristics}
\label{subsec:dataset_char}
SDV-W follows the high-resolution rendering configuration of SDV, with images provided at a fixed resolution of 2560$\times$1477 pixels. Drone instances exhibit broad image-plane coverage, yielding a spatially diverse object-location distribution. Object area ratios range from 1$\times$10$^{-5}$ to 0.761 (mean:~0.0075), indicating a systematic shift toward smaller projected object extents relative to SDV. Object aspect ratios range from 0.205 to 8.222 (mean:~1.243), suggesting that the selected sequences preserve the orientation and observation-angle variability of the original dataset. Following the COCO taxonomy at the default YOLO input resolution of 640$\times$640, SDV-W is primarily composed of medium (57.0\%) and small (30.9\%) objects, with fewer large (10.4\%) and x-small (1.6\%) instances (Fig.~\ref{fig:Dist_ObjectSizeCat_SDVW}).

\section{Experimental Setup}
\label{sec:exp_setup}
This section details the experimental setup for assessing the effect of weather-conditioned synthetic data on real-world drone detection.

\begin{table}[t!]
  \caption{\label{tab:SDV_Weather_Stats} Composition of SDV-W across seasons and weather.\vspace{-0.2cm}}
  \centering\footnotesize
  \begin{tabular}{@{}|l|c|c|c|c|@{}}
    \hline
    Weather Cond.  & \multicolumn{3}{c|}{Season} & Total\\\cline{2-4}
    & spring/summer & autumn & winter & \\\hline
    rain   & 5,598 & 5,200 & 5,199 & 15,997\\\hline
    fog    & 4,400 & 5,200 & 5,197 & 14,797\\\hline
    rain$+$fog & 2,396 & 2,600 & 2,600 & 7,596\\\hline
    sunny  & 200 & 2,000 & 2,600 & 4,800\\\hline
    snow   & -- & -- & 6,000 & 6,000\\\hline
    leaves & -- & 5,997 & -- & 5,997 \\\hline
    \textbf{total}  & \textbf{12,594} & \textbf{20,997} & \textbf{21,596} & \textbf{55,187}\\\hline
  \end{tabular}
\end{table}

\subsection{Models}
\label{subsec:models}
The selected detectors cover representative drone detection architectures, enabling a separation of training-data effects from architecture-dependent performance variation. The analysis considers two complementary categories of detection model. The first comprises generic object detection models widely adopted in drone detection applications. Specifically, it includes standard YOLO configurations spanning successive architectural generations and capacity scales: YOLOv5l~\cite{Ultralytics_v5}, YOLOv8m/l~\cite{Varghese:2024,Ultralytics_v8}, YOLOv9c/e~\cite{Wang:2024}, and YOLOv11l/x~\cite{Ultralytics_v11}. The second category comprises detectors developed specifically for urban drone detection and is represented here by YOLO-FEDER FusionNet~\cite{Lenhard:2024,Lenhard:2026}. YOLO-FEDER FusionNet combines generic object detection with camouflage-aware detection principles, targeting camouflage-induced target-background ambiguity in urban scenes (\eg, drones blending with vegetation or built structures) -- conditions under which standard YOLO models are prone to failure~\cite{Lenhard:2024}. 

\begin{figure*}[t!]
\centering
  \includegraphics[width=\textwidth, trim={0.2cm 26.1cm 0.2cm 0.cm}, clip]{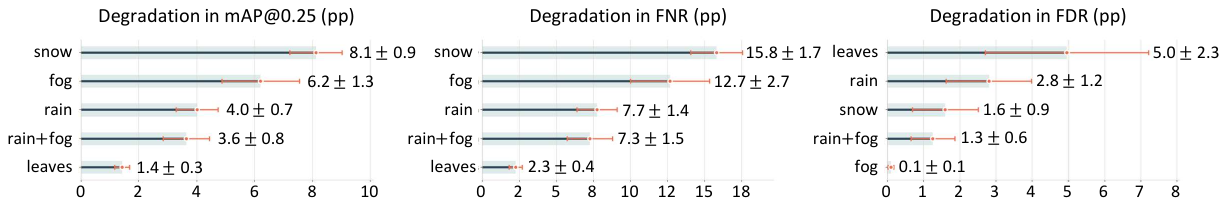}
  \caption{\label{fig:PerfomanceDrop_Weather_1}Weather-induced degradation in YOLO detection performance relative to clear conditions. Values give the magnitude of performance loss in percentage points (pp), defined as decreased mAP@0.25 or increased FNR/FDR. Larger values indicate greater degradation. Orange markers show the mean across YOLO models. Horizontal intervals and $\pm$ annotations represent inter-model standard deviation.}
\end{figure*}

\subsection{Data}
\label{subsec:data}
The experimental setup combines synthetic and real-world datasets for drone detection. The synthetic basis comprises SDV~\cite{Lenhard:2025} and its weather-conditioned extension SDV-W, providing large-scale visual data with controlled seasonal and adverse-weather variation. 

Complementing this controlled setting, the real-world basis comprises publicly available datasets and custom recordings spanning diverse urban environments, illumination conditions, and seasonal appearances. Here, DUT Anti-UAV~\cite{Zhao:2022} serves as a general-purpose urban drone detection dataset featuring heterogeneous backgrounds, varying lighting conditions, and diverse drone models. LRDDv1~\cite{Rouhi:2024} and LRDDv2~\cite{Rouhi:2025}\footnote{The more recent LRDDv3 dataset~\cite{Peterson:2026}, available upon request, could not be obtained in time for the experiments.} extend coverage toward longer-range urban scenarios and, unlike DUT Anti-UAV, capture seasonal appearance variation through summer and winter recordings with distinct vegetation states. Additionally, two custom recordings, R1-POS3 and R2-POS7, are included to provide controlled real-world seasonal variation (Fig.~\ref{figExamples_custom}, Supp.). Both were acquired in urban environments under summer and winter conditions. By preserving camera position, \textit{field of view (FOV)}, and scene geometry across seasons, they isolate seasonal appearance changes from scene-content variation, thereby constituting a controlled real-world counterpart to SDV and SDV-W. The selected datasets exhibit pronounced differences in object-size distribution and effective target scale (Fig.~\ref{fig:Dist_ObjectSizeCat_Datasets_640x640}, Supp.). At an input resolution of 640$\times$640 pixels, effective object scales vary from the comparatively balanced distribution in DUT Anti-UAV to settings dominated by small-scale instances (\eg, LRDDv2 and R2-POS7). Detailed statistics are provided in Sec.~D (Supp.).

\subsection{Training Protocol}
\label{subsec:training_protocol}
Model training is performed under two primary data configurations. The reference setting follows~\cite{Lenhard:2025} and combines large-scale synthetic data from SDV with real-world samples from DUT Anti-UAV~\cite{Zhao:2022}. The combined training set comprises 140,038 SDV images and 7,800 DUT Anti-UAV images from the respective training and validation splits, yielding a real-data share of $\sim$5\%. The other configuration incorporates SDV-W alongside SDV and DUT Anti-UAV, yielding 188,865 training and 14,162 validation images. Thus, SDV-W is integrated additively by default, with deviations specified explicitly. All models are trained (and evaluated) at a common input resolution of 640$\times$640 pixels, with images resized accordingly. While standard YOLO detectors use letterbox resizing (i.e., aspect-ratio-preserving scaling with padding), YOLO-FEDER FusionNet applies dynamic shortest-side cropping followed by resizing to obtain square, padding-free inputs~\cite{Lenhard:2026}.

To ensure comparability across training data compositions, all detectors are optimized under a fixed training protocol. Models are trained for 100 epochs with a batch size of 64 on four NVIDIA A100 GPUs, using default Ultralytics hyperparameters unless stated otherwise~\cite{Ultralytics_v5,Ultralytics_v8,Ultralytics_v11}. Standard YOLO detectors use COCO-pretrained initialization~\cite{Lin:2014}. YOLO-FEDER FusionNet follows~\cite{Lenhard:2026}, initializing its YOLOv8l backbone from COCO~\cite{Lin:2014} and its FEDER~\cite{He:2023} branch from COD10K~\cite{Fan:2020} pretrained weights. Both branches remain frozen, restricting training to the fusion neck and detection head.

\subsection{Evaluation Procedure}
\label{subsec:eval_protocol}
Evaluation is conducted on the public datasets DUT Anti-UAV, LRDDv1, and LRDDv2, as well as on the custom recordings R1-POS3 and R2-POS7. Detection performance is quantified using standard object detection metrics: \textit{mean average precision (mAP)} at an \textit{intersection over union (IoU)} threshold of 0.5, mAP averaged over IoU thresholds from 0.5 to 0.95, and mAP at a relaxed IoU threshold of 0.25 to account for annotation-induced localization uncertainty~\cite{Lenhard:2024}. To capture operational failure modes, we additionally report the \textit{false negative rate (FNR)} and \textit{false discovery rate (FDR)}, reflecting missed detections and false alarms. Given the strong dependence of drone detection performance on effective target scale and the heterogeneous object-size distributions across datasets, performance is also evaluated across COCO-style object-size categories (Tab.~\ref{tab:obj_size_categories}, Supp.). Departing from COCO's native-image size definition, object-size categories are assigned based on the ground-truth bounding box areas after resizing to 640$\times$640 (input resolution). This aligns the taxonomy with the detector's operating resolution and accounts for resizing-induced category transitions (Tab.~\ref{tab:size_bucket_distribution}, Supp.).

\section{Results}
\label{sec:results}
The evaluation first leverages SDV-W's controlled design to quantify weather-induced detector degradation. It then assesses SDV-W's primary role as weather-conditioned training resource real-world detection, disentangling the effects of data diversity, dataset scale, and target size.

\begin{table*}[t!]
  \caption{\label{tab:results_YOLOs_DUT_and_LRDDv1}Comparison of training with and without weather-condition synthetic data (SDV-W) on LRDDv1 and DUT Anti-UAV across multiple YOLO architectures and model scales. Best-performing values are highlighted in \textbf{bold}. \ding{51} indicates training with SDV-W.\vspace{-0.2cm}}
  \centering\footnotesize
  \begin{tabular}{@{}|l|c|ccc|c|c|ccc|c|c|@{}}
  
    \multicolumn{2}{c}{} & \multicolumn{5}{c}{LRDDv1} & \multicolumn{5}{c}{DUT Anti-UAV} \\
    \hline
    Model & w/ & \multicolumn{3}{c|}{mAP~$\uparrow$} & FNR~$\downarrow$ & FDR~$\downarrow$ &  \multicolumn{3}{c|}{mAP~$\uparrow$} & FNR~$\downarrow$ & FDR~$\downarrow$\\
    & SDV-W & @0.25 & @0.5 & @0.5-0.95 & & & @0.25 & @0.5 & @0.5-0.95 & &\\\hline
    \multirow{2}{*}{YOLOv5l} 
    
    & -- 
    & 0.580 & 0.426 & 0.131 & 0.489 & 0.077
    & \textbf{0.957} & \textbf{0.937} & 0.674 & \textbf{0.102} & 0.035 
    \\ 

    & \ding{51} 
    & \textbf{0.700} & \textbf{0.566} & \textbf{0.204} & \textbf{0.486} & \textbf{0.030}
    & 0.936 & 0.928 & \textbf{0.723} & 0.108 & \textbf{0.031} 
    \\\hline
    
    \multirow{2}{*}{YOLOv8m} 
    & -- 
    & 0.606 & 0.442& 0.135 & 0.480 & 0.059
    & 0.954 & 0.932 & 0.680 & 0.106 & \textbf{0.033} 
    \\
    & \ding{51} 
    & \textbf{0.711} & \textbf{0.583} & \textbf{0.214} & \textbf{0.474} & \textbf{0.023}
    & \textbf{0.963} & \textbf{0.945} & \textbf{0.691} & \textbf{0.053} & 0.076 
    \\\hline
    
    \multirow{2}{*}{YOLOv8l} 
    & -- & 0.608 & 0.430 & 0.132 & 0.470 & 0.052 & 0.960 & 0.940 & 0.692 & 0.096 & \textbf{0.027} 
    \\	
    & \ding{51} & \textbf{0.700} & \textbf{0.553} & \textbf{0.197} & \textbf{0.468} & \textbf{0.025} & \textbf{0.967} & \textbf{0.953} & \textbf{0.694} & \textbf{0.051} & 0.046 
    \\\hline
    
    \multirow{2}{*}{YOLOv9c} 
    & -- 
    & 0.614 & 0.442 & 0.139 & 0.484 & 0.043
    & 0.960 & 0.939 & 0.690 & 0.097 & \textbf{0.027} 
    \\	
    & \ding{51} 
    & \textbf{0.714} & \textbf{0.569} & \textbf{0.202} & \textbf{0.452} & \textbf{0.024}
    & \textbf{0.969} & \textbf{0.953} & \textbf{0.696} & \textbf{0.040} & 0.058 
    \\\hline	
    
    \multirow{2}{*}{YOLOv9e} 
    & -- 
    & 0.632 & 0.440 & 0.134 & \textbf{0.455} & 0.016
    & 0.970 & 0.954 & \textbf{0.722} & 0.094 & \textbf{0.007}
    \\	
    & \ding{51} 
    & \textbf{0.726} & \textbf{0.573} & \textbf{0.207} & 0.464 & \textbf{0.012}
    & \textbf{0.976} & \textbf{0.965} & 0.713 & \textbf{0.037} & 0.032 
    \\\hline	
    
    \multirow{2}{*}{YOLOv11l} 
    & -- 
    & 0.634 & 0.463 & 0.145 & 0.465 & \textbf{0.018}
    & 0.963 & 0.942 & 0.697 & 0.100 & \textbf{0.026} 
    \\	
    & \ding{51} 
    & \textbf{0.710} & \textbf{0.567} & \textbf{0.207} & \textbf{0.464} & 0.022
    & \textbf{0.970} & \textbf{0.959} & \textbf{0.705} & \textbf{0.042} & 0.042 
    \\\hline	
    
    \multirow{2}{*}{YOLOv11x} 
    & -- 
    & 0.617 & 0.446 & 0.142 & \textbf{0.469} & \textbf{0.012}
    & 0.966 & 0.944 & \textbf{0.708} & 0.091 & \textbf{0.028} 
    \\	
    & \ding{51} 
    & \textbf{0.700} & \textbf{0.569} & \textbf{0.210} & 0.477 & 0.023
    & \textbf{0.971} & \textbf{0.960} & 0.701 & \textbf{0.046} & 0.034 
    \\\hline
  \end{tabular}
\end{table*}

\begin{table*}[t!]
  \caption{\label{tab:results_YOLO-FEDER_R1_R2}Performance of YOLO-FEDER FusionNet with and without weather-conditioned synthetic training data (SDV-W) under seasonal conditions (summer vs. winter) on the datasets R1-POS3 and R2-POS7. Results are aggregated across all size categories. Size-specific performance indicators are provided in the Supp. Best-performing values are highlighted in \textbf{bold}. \ding{51} indicates training with SDV-W.\vspace{-0.2cm}}
  \centering\footnotesize
  \begin{tabular}{@{}|l|c|ccc|c|c|ccc|c|c|@{}}
     \multicolumn{2}{c}{} & \multicolumn{5}{c}{Summer} & \multicolumn{5}{c}{Winter}\\\hline
    Dataset & w/ &  \multicolumn{3}{c|}{mAP~$\uparrow$} & FNR~$\downarrow$ & FDR~$\downarrow$ & \multicolumn{3}{c|}{mAP~$\uparrow$} & FNR~$\downarrow$ & FDR~$\downarrow$\\
    & SDV-W & @0.25 & @0.5 & @0.5-0.95 & & & @0.25 & @0.5 & @0.5-0.95 & &\\\hline
    
    \multirow{2}{*}{R1-POS3}
    & -- 
    & 0.764 & 0.743 & 0.331 & \textbf{0.313} & 0.083
    & 0.931 & 0.827 & \textbf{0.404} & 0.208 & 0.006\\
    & \ding{51} 
    & \textbf{0.821} & \textbf{0.794} & \textbf{0.354} & 0.326 & \textbf{0.004} 
    & \textbf{0.965} & \textbf{0.836} & 0.393 & \textbf{0.164} & \textbf{0.004}\\\hline
    
    \multirow{2}{*}{R2-POS7}
    & -- 
    & \textbf{0.985} & \textbf{0.909} & \textbf{0.450} & 0.076 & 0.010
    & 0.983 & \textbf{0.749} & \textbf{0.294} &  \textbf{0.025} & 0.054\\
    & \ding{51} 
    & 0.982 & 0.901 & 0.422 & \textbf{0.061} & \textbf{0.008}
    & \textbf{0.988} & 0.675 & 0.223 & 0.028 & \textbf{0.006}\\\hline

  \end{tabular}
\end{table*}

\subsection{Performance under Synthetic Weather \& Seasonal Variation}
\label{subsec:degradation_SDV-W}
To quantify the impact of atmospheric appearance shifts on detector performance, standard YOLO models trained on a combination of SDV and DUT Anti-UAV are evaluated on SDV-W and their corresponding clear-weather SDV counterparts. By pairing adverse-weather samples with identical SDV scenes under clear conditions, \mbox{SDV-W} provides a controlled counterfactual for attributing performance gaps to weather-induced appearance changes rather than scene-content variation. Averaged across YOLO configurations, all adverse conditions reduce performance relative to their clear-weather counterparts, with  an architecture-stable severity ordering (Fig.~\ref{fig:PerfomanceDrop_Weather_1}). Snow and fog are most disruptive, reducing mAP@0.25 by 8.1/6.2\,pp and raising FNR by 15.8/12.7\,pp , respectively. Under stricter IoU criteria, this degradation becomes more pronounced (Fig.~\ref{fig:PerfomanceDrop_Weather}, Supp.): for snow, the drop increases from 8.2\,pp  at mAP@0.25 to 12.2\,pp  at mAP@0.5-0.95, indicating an additional localization penalty. Rain-based conditions induce only moderate mAP degradation, while falling leaves have negligible impact on overall accuracy (Fig.~\ref{fig:PerfomanceDrop_Weather_1}, left). 

At the error-type level, the dominant failure mode is missed detection. Across conditions, FNR increases substantially more than FDR (Fig.~\ref{fig:PerfomanceDrop_Weather_1}, middle/right). For instance, under snow and fog, the FDR remains nearly unchanged despite the largest FNR increases. Thus, adverse weather appears to suppress object saliency below the detection threshold rather than induce \textit{false positives (FPs)}. The only notable exception is falling leaves, where added visual noise increases FDR despite limited impact on mAP.

\subsection{Impact of SDV-W-based Training}
\label{subsec:res_training_impact}
The contribution of SDV-W as a training resource is quantified by contrasting the reference data composition with its SDV-W-extended counterpart across detectors and datasets.

\vspace{-0.3cm}
\paragraph{Performance on Public Datasets.} On public real-world datasets, SDV-W provides the clearest benefits in settings where reference performance is not already saturated. The most consistent gains are observed on LRDDv1: across all seven standard YOLO variants, mAP@0.5 increases by $+$10.4 to $+$14.1\,pp (mean:~$+$12.7\,pp), with simultaneous gains in mAP@0.25 and mAP@0.5-0.95 (Tab.~\ref{tab:results_YOLOs_DUT_and_LRDDv1}). As FNR remains largely unchanged, the gains mainly reflect improved localization and confidence calibration, rather than increased target recovery. On DUT Anti-UAV, where reference performance is already close to saturation, \mbox{SDV-W} yields only marginal mAP improvements, yet lowers FNR by 4.4\,pp on average. This indicates fewer missed detections without compromising reference-domain performance. YOLO-FEDER FusionNet follows the same saturation pattern (Tab.~\ref{tab:results_YOLO-FEDER_PublicData}, Supp.). On LRDDv2, the effect of SDV-W is less uniform: mAP changes are ar\-chitecture-dependent and approximately neutral on average, while FDR decreases consistently. Given the high FNR across training configurations ($\geq$0.61; Tab.~\ref{tab:results_YOLOs_LRDD_v2}, Supp.), mAP has limited interpretive value. In contrast, YOLO-FEDER FusionNet achieves substantially lower FNR on LRDDv2 than any standard YOLO variant (Tab.~\ref{tab:results_YOLO-FEDER_LRDDv2}) and benefits more consistently from SDV-W. Across all IoU criteria, SDV-W improves mAP while reducing both FNR and FDR. The largest gains occur for medium objects (mAP@0.5: $+$9.0\,pp, FNR: $-$10.4\,pp), with consistent improvements for small and large objects. Extra-small objects remain challenging.

\begin{table}[t!]
  \caption{\label{tab:results_YOLO-FEDER_LRDDv2} Impact of weather-conditioned synthetic data (SDV-W) on the performance of YOLO-FEDER FusionNet on LRDDv2, reported overall and by size category. Best-performing values are highlighted in \textbf{bold}. \ding{51} indicates training with SDV-W.\vspace{-0.2cm}}
  \centering\footnotesize
  \begin{tabular}{@{}|c|c|ccc|c|c|@{}}
    \hline
    w/ & Size & \multicolumn{3}{c|}{mAP~$\uparrow$} & FNR~$\downarrow$ & FDR~$\downarrow$\\
    SDV-W & Cat. & @0.25 & @0.5 & @0.5-0.95 & &\\\hline

    -- & \multirow{2}{*}{all} & 0.540 & 0.507 & 0.254 & 0.515 & 0.270\\
    \ding{51} & & \textbf{0.561} & \textbf{0.530} & \textbf{0.258} & \textbf{0.491} & \textbf{0.235}\\\hline
    
    -- & \multirow{2}{*}{XS} & 0.144 & 0.125 & 0.052 & 0.743 & --\\
    \ding{51} & & \textbf{0.152} & \textbf{0.133} & \textbf{0.055} & \textbf{0.735} & -- \\\hline
    
    -- & \multirow{2}{*}{S} & 0.683 & 0.668 & \textbf{0.314} & 0.253 & --\\
    \ding{51} & & \textbf{0.705} & \textbf{0.690} & 0.311 & \textbf{0.227} & -- \\\hline
    
    -- & \multirow{2}{*}{M} & 0.669 & 0.665 & 0.399 & 0.242 & --\\
    \ding{51} & & \textbf{0.762} & \textbf{0.755} & \textbf{0.448} & \textbf{0.138} & -- \\\hline
    
    -- & \multirow{2}{*}{L} & 0.827 & 0.826 & 0.587 & 0.053 & --\\
    \ding{51} & & \textbf{0.856} & \textbf{0.855} & \textbf{0.600} & \textbf{0.031} & -- \\\hline
  \end{tabular}
\end{table}

\vspace{-0.3cm}
\paragraph{Performance on Custom Seasonal Recordings.} Evaluation on R1-POS3 and R2-POS7 is confined to YOLO-FEDER FusionNet, as tree-covered backgrounds induce camouflage effects known to reduce the reliability of generic detection models~\cite{Lenhard:2024}. On R1-POS3, where baseline performance remains below saturation, \mbox{SDV-W} yields consistent improvements across both seasons (Tab.~\ref{tab:results_YOLO-FEDER_R1_R2}). In summer, mAP@0.5 increases by $+$5.1\,pp, while FDR drops by over an order of magnitude at nearly unchanged FNR. In winter, mAP@0.5 improves modestly, while FNR decreases by 4.4\,pp. Consistent with LRDDv2, remaining limitations are concentrated among extra-small objects. Performance on R2-POS7 is near saturation, limiting the measurable effect of SDV-W, particularly under summer conditions. Under winter conditions, SDV-W markedly reduces FDR at unchanged FNR. The mAP drop at stricter IoU thresholds mainly reflects reduced localization accuracy for small objects, as performance for medium objects improves markedly (Tab.~\ref{tab:results_YOLO-FEDER_R1_R2_2}, Supp.).

\subsection{Weather Diversity vs. Training-Set Size}
\label{subsec:res_ablation}
To isolate the effect of SDV-W-induced weather diversity from increased training volume, we replace a randomly selected, size-matched subset of SDV samples with SDV-W while keeping the total training-set size fixed. YOLO-FEDER FusionNet is trained on five independently sampled training variants, and performance is reported as the per-dataset mean across runs (Tab.~\ref{tab:ablation_study}, Supp.). Across datasets, the replacement yields the most pronounced average gains on the custom recordings, where season-matched samples appear to better align the training distribution with the fixed-FOV target domain. On the larger and more diverse public datasets, replacing general-purpose SDV samples with SDV-W provides no consistent benefit, suggesting that the added weather variation does not compensate the loss of broader synthetic coverage. Collectively, this pattern indicates that SDV-W is most effective when used to complement, rather than replace, general-purpose synthetic training data. Details are provided in Sec.~\ref{subsec:ablation} (Supp.).

\subsection{Resolution-Dependent Detection Trade-offs}
\label{subsec:res_tiling}
Extra-small objects remain the dominant residual failure mode across datasets despite SDV-W-based training, as shown in Tab.~\ref{tab:results_YOLO-FEDER_LRDDv2}. To investigate the contribution of input resolution to this limitation, conventional resizing to 640$\times$640 is replaced with tiled inference. Tiled inference preserves higher effective object resolution by independently processing overlapping 640$\times$640 patches, followed by prediction fusion via SAHI-style \textit{greedy non-maximum merging} with \textit{intersection over smaller area} matching~\cite{Akyon:2022}. Applying tiled inference to the SDV-W-trained YOLO-FEDER FusionNet substantially improves extra-small object detection. On LRDDv2, mAP@0.5 more than doubles (0.133$\rightarrow$0.312), while FNR decreases substantially (0.735$\rightarrow$0.434). Performance for small objects remains largely unchanged. However, these gains come with clear trade-offs: large-object localization deteriorates (mAP@0.5:~0.855$\rightarrow$0.661) due to tile-boundary fragmentation, while image-level FDR increases markedly (0.235$\rightarrow$0.426). Thus, improved small-object recovery comes at the expense of large-object fidelity, precision, and computational efficiency, with inference time increasing from 12 to $>$50\,ms (Sec.~\ref{subsec:tiling_supp}, Supp.).

\section{Discussion \& Conclusion}
\label{sec:conclusion}
The analyses indicate that SDV-W extends operational-domain coverage for urban drone detection along the weather$\times$season axis through controlled environmental variation. Beyond its role as a training data source, its paired imagery and graded severity levels enable systematic, condition-specific assessment of detector degradation, rarely feasible with existing real-world datasets. On real-world data, its benefits are primarily reflected in improved detection reliability, with fewer missed detections and false alarms -- both critical for operational deployment -- while localization gains are largely confined to non-saturated baselines. Overall, SDV-W is most effective as a complementary training source, broadening environmental variation without sacrificing cross-domain generalization potential -- at least for CNN-based detectors. While SDV-W mitigates appearance-driven degradation, residual detection failures remain predominantly resolution-driven. Tiled inference improves sensitivity at this scale but compromises large-object accuracy, precision, and computational efficiency. Given the distance-dependent scale variation in drone surveillance, these trade-offs motivate future work on selective, scale-adaptive inference strategies.

Nevertheless, a key limitation remains the limited availability of real-world drone data under diverse adverse weather conditions (see Sec.~\ref{sec:scope_and_limitations}, for additional limitations). Collecting such data to further validate the benefits of SDV-W remains an important direction for future work.

{
    \small
    \bibliographystyle{ieeenat_fullname}
    \bibliography{main}
}

\newpage
\twocolumn[{%
 \centering
 \vspace{1cm}
  \Large \textbf{Supplementary Material} \\[5em]
}]

\appendix
\setcounter{table}{0}
\setcounter{figure}{0}
\renewcommand{\thetable}{\Roman{table}}
\renewcommand{\thefigure}{\Roman{figure}}

\appendix

\begin{figure}[t!]
\centering
  \includegraphics[width=0.49\textwidth, trim={0.1cm 18cm 10.1cm 0cm}, clip]{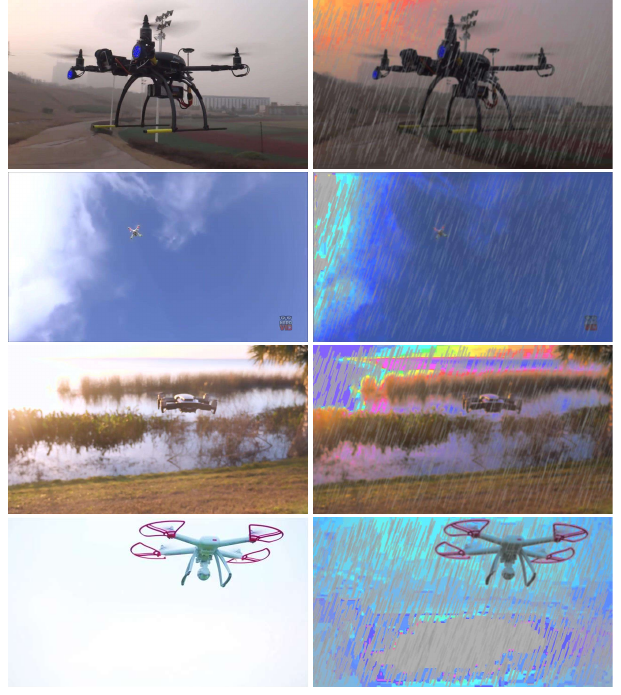}
  \caption{\label{fig:RW_ExamplesOtherData} Example of image-space rain augmentation. \textbf{Left:} clear reference frames from DUT Anti-UAV~\cite{Zhao:2022} (Apache-2.0). \textbf{Right:} corresponding frames from RTS~\cite{Munir:2023,Munir:2024}. Rain streaks are superimposed uniformly in image space, irrespective of scene depth, while original illumination cues (\eg, direct sunlight and cast shadows) remain unchanged. The images further exhibit spatially nonuniform chromatic distortions, including artifactual color shifts in sky (rows~1-2) and water regions (row~3), and locally homogeneous gray areas (row~4).}
\end{figure}

\section{Image-Space Rain Augmentation}
Fig.~\ref{fig:RW_ExamplesOtherData} contrasts clear DUT Anti-UAV frames with their rain-augmented counterparts from RTS~\cite{Munir:2023,Munir:2024}, highlighting the limitations of image-space augmentation discussed in Sec.~\ref{subsec:RW_data} (main paper). As rain is superimposed as a 2D overlay, streak appearance lacks depth dependence, while scene illumination remains unchanged. Moreover, the augmentation introduces color artifacts uncharacteristic of real rainy scenes. In contrast, simulation-based rendering, as employed in SDV-W, mitigates these limitations by modeling precipitation directly within the 3D scene, accounting for scene geometry and lighting.

\section{Additional Details on SDV-W}
This section provides additional details on the generation of SDV-W, including sequence selection, frame sampling, as well as weather- and season-conditioned rendering.

\subsection{Sequence Selection \& Frame Sampling}
\begin{figure}[t!]
\centering
  \includegraphics[width=0.47\textwidth, trim={0.1cm 17.8cm 10cm 0cm}, clip]{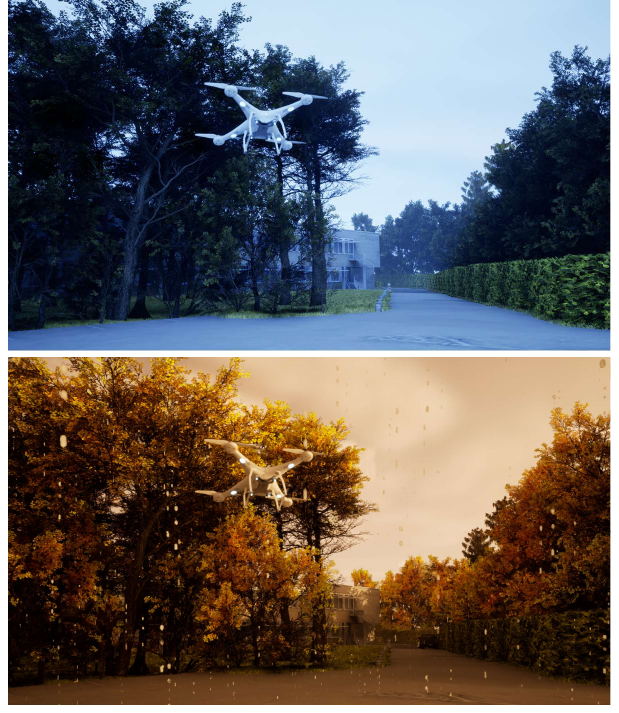}
  \caption{\label{fig:supp_example_pairing} Paired SDV/SDV-W rendering example. The original SDV frame is shown above, and the corresponding SDV-W rendering under rainy autumn conditions is shown below.}
\end{figure}

SDV-W is derived from a curated subset of 15 source sequences selected from the 72-sequence \textit{SynDroneVision (SDV)} dataset~\cite{Lenhard:2025}. The selected sequence identifiers and corresponding environments are listed in Tab.~\ref{tab:sdvw_sequence_selection}. For frame sampling, each selected sequence was considered in its entirety, independent of the original SDV train/validation/test split assignment. Preserving the SDV sequence structure yields paired clean-condition frames and corresponding seasonal or adverse-weather counterparts, with examples shown in Fig.~\ref{fig:supp_example_pairing}.

\begin{table}[t!]
\centering\footnotesize
\caption{\label{tab:sdvw_sequence_selection}
SDV source sequences underlying the generation of SDV-W, grouped by their corresponding environments.\vspace{-0.1cm}}
\begin{tabular}{|C{2.2cm}|C{2.2cm}|C{2.4cm}|}
\hline
\multicolumn{3}{|c|}{Environment}\\\hline
University Site & Venetian City~\cite{Unreal:Venice} & Urban Downtown~\cite{Unreal:DowntownWest}\\\hline
seq0027 & seq0009 & seq0001\\
seq0062 & seq0011 & seq0047\\
seq0063 & seq0014 & seq0048\\
seq0070 & seq0016 & seq0051\\
seq0072 & seq0017 & \\
& seq0018 & \\\hline
\end{tabular}
\end{table}

\begin{table*}[t!]
\centering\footnotesize
\caption{\label{tab:material_recoloring_downtown}
Material-level foliage recoloring parameters for the Urban Downtown~\cite{Unreal:DowntownWest} autumn configuration.\vspace{-0.1cm}}
\begin{tabular}{|l|c|c|c|c|c|c|}
\hline
Material & \multicolumn{4}{c|}{\texttt{Albedo\_1\_ColorOverlay}} & \texttt{Albedo\_Brightness} & \texttt{Albedo\_Add} \\
& R & G & B & A & &\\\hline
\texttt{MI\_Foliage\_Tree\_Pine\_Leaves} & 0.617 & 0.255 & 0.000 & \ \ \ 9.000 & 1.756 & \ \ \ 0.198 \\\hline
\texttt{MI\_Tree\_Narrowleaf\_Leaf\_A} & 1.000 & 0.785 & 0.000 & $-$0.091 & 1.522& $-$0.581\\\hline
\texttt{MI\_Foliage\_Trre\_Leaves} & 0.976 & 0.718 & 0.062 & \ \ \ 0.000 & -- & \ \ \ -- \\\hline
\end{tabular}
\end{table*}

\subsection{Technical Details on Seasonal Variations}
SDV-W parameterizes seasonal appearance primarily through vegetation, the dominant visual cue in mid-latitude urban environments with pronounced canopy variation. Three vegetation states are instantiated: dense green foliage for spring/summer, yellow-brown foliage for autumn, and defoliated or snow-covered trees for winter. The implementation is environment-specific, combining native foliage presets, material-level color modulation, and season-specific asset replacement according to the available vegetation assets.

\vspace{-0.3cm}
\paragraph{University Site.} The University Site uses tree assets from~\cite{Unreal:BlackAlder}. The spring/summer state corresponds to the default green-foliage configuration, while winter uses the native winter variant. Autumn is realized through modifications of the \texttt{BlackAlder\_TwoSided} leaf material, setting \texttt{Albedo Tint Leaves} to $(\text{R},\text{G},\text{B})\!=\!(1.939,0.810,0.384)$ and the tint strength to 8.419.

\vspace{-0.3cm}
\paragraph{Venetian City.} The Venetian City environment~\cite{Unreal:Venice} includes vegetation assets with native seasonal color variants.  The spring/summer configuration uses green foliage, while the autumn configuration combines yellow and orange variants. Since no defoliated vegetation variant is available by default, the winter configuration is implemented by replacing the native vegetation with assets from~\cite{Unreal:WinterForest}.

\vspace{-0.3cm}
\paragraph{Urban Downtown.} The Urban Downtown environment \cite{Unreal:DowntownWest} provides vegetation only in a green-foliage configuration. Winter is realized by replacing the native vegetation with assets from~\cite{Unreal:WinterForest}, while autumn is obtained through material-level recoloring. The modified material parameters are listed in Tab.~\ref{tab:material_recoloring_downtown}.

\subsection{Illumination Parameterization}
\label{subsec:illumination_params}
Season- and weather-specific appearance variations are realized through environment-specific UE5 lighting actors and post-processing parameters governing solar position, direct illumination, atmospheric scattering, ambient sky illumination, and global color grading. 

For the University Site and Venetian City~\cite{Unreal:Venice} environments, these variations are implemented via the Sun and Sky Actor~\cite{Unreal:SunSkyActor}, parameterizing solar time and selected attributes of the associated Directional Light~\cite{Unreal:DirectionalLight}, Sky Atmosphere~\cite{Unreal:SkyAtmosphere}, and Sky Light~\cite{Unreal:SkyLight} components. Tab.~\ref{tab:SDVW_Illumination_Params_Dieburg_Venice}  reports the corresponding solar-time intervals, Directional Light intensity ranges, and channel-wise Rayleigh scattering ranges. Directional-light and skylight colors remain near-neutral, with approximately balanced RGB channels, and vary primarily in luminance from dark overcast gray with RGB values of $\sim$80 to white. Chromatic deviations are limited to warm pale-yellow tones, approximately (255, 255, 225), and cool blue/blue-gray casts, approximately (160, 182, 255) and (167, 183, 202), in selected autumn-rain and winter-fog configurations. For the Urban Downtown~\cite{Unreal:DowntownWest} environment, the native Directional Light~\cite{Unreal:DirectionalLight} and SkyLight~\cite{Unreal:SkyLight} actors define the lighting model. Tab.~\ref{tab:DirLight_Downtown} reports the Directional Light intensity and azimuthal rotation (Z). The Directional Light's elevation parameters are held at their default values, X=0$^\circ$ and Y=-46$^\circ$. Directional Light and Sky Light colors are primarily neutral white-to-gray, with reduced RGB magnitudes of approximately 160 under overcast and snow conditions.

\begin{table*}[t!]
\centering\footnotesize
  \caption{\label{tab:SDVW_Illumination_Params_Dieburg_Venice} Sun and Sky Actor~\cite{Unreal:SunSkyActor} parameter ranges for SDV-W lighting variation across University Site and Venetian City~\cite{Unreal:Venice}, aggregated over weather severity levels.\vspace{-0.1cm}}
  \begin{tabular}{|l|l|c|c|C{1.1cm}|C{1.1cm}|c|c|c|c|c|c|c|}
    \hline
    Season & Weather & \multicolumn{2}{c|}{Solar Time} & \multicolumn{2}{c|}{Direct. Light Intensity} & \multicolumn{6}{c|}{Rayleigh Scattering (Channel Values)}\\
    & Cond. & \multicolumn{2}{c|}{} & \multicolumn{2}{c|}{(lux)} & \multicolumn{2}{c|}{R} & \multicolumn{2}{c|}{G} & \multicolumn{2}{c|}{B}\\
    & & from & to & min & max & min & max & min & max & min & max\\\hline
    \multirow{4}{*}{spring/summer} 
    & sunny    & 11.44 & 12.19 & 6.4 & 10.0 & 0.175 & 0.175 & 0.410 & 0.410 & 1.000 & 1.000\\
    & rain     & 9.57 & 20.13  & 0.5 & 9.0  & 0.000 & 0.806 & 0.192 & 0.861 & 0.212 & 1.000\\
    & fog      & 6.59 & 11.64  & 1.0 & 8.0 & 0.028 & 1.000 & 0.047 & 0.834 & 0.553 & 1.000\\
    & rain$+$fog & 9.57 & 20.13  & 0.5 & 9.0  & 0.175 & 0.806 & 0.192 & 0.861 & 0.212 & 1.000\\\hline
    
    \multirow{5}{*}{autumn}  
    & sunny    & 9.68 & 18.22  & 1.0 & 6.0  & 0.014 & 0.656 & 0.204 & 0.629 & 0.073 & 1.000\\
    & rain     & 8.38 & 18.69  & 1.0 & 16.6 & 0.014 & 0.876 & 0.204 & 0.884 & 0.073 & 1.000\\
    & fog      & 8.50 & 11.53  & 0.5 & 16.6 & 0.147 & 1.000 & 0.137 & 0.848 & 0.264 & 1.000\\
    & rain$+$fog & 8.38 & 18.50  & 1.0 & 16.6 & 0.014 & 0.876 & 0.204 & 0.884 & 0.073 & 1.000\\
    & leaves   & 10.28 & 16.41 & 0.5 & 6.4  & 0.000 & 0.543 & 0.410 & 0.851 & 0.983 & 1.000\\\hline
    
    \multirow{5}{*}{winter}  
    & sunny    & 10.00 & 16.00 & 2.0 & 10.0 & 0.011 & 0.175 & 0.242 & 0.719 & 0.535 & 1.000\\
    & rain     & 8.43 & 17.01  & 2.0 & 10.0 & 0.007 & 0.893 & 0.007 & 0.948 & 0.007 & 1.000\\
    & fog      & 7.32 & 16.00  & 1.0 & 12.0 & 0.031 & 0.664 & 0.137 & 0.719 & 0.778 & 1.000\\
    & rain$+$fog & 8.43 & 17.01  & 2.0 & 10.0 & 0.007 & 0.893 & 0.007 & 0.948 & 0.007 & 1.000\\
    & snow     & 8.00 & 16.90  & 0.5 & 5.0  & 0.000 & 0.979 & 0.000 & 0.985 & 0.000 & 1.000\\\hline
  \end{tabular}
\end{table*}

\begin{table}[t]
\centering\footnotesize
\caption{\label{tab:DirLight_Downtown}Directional Light~\cite{Unreal:DirectionalLight} parameter ranges for Urban Downtown~\cite{Unreal:DowntownWest}. Intensity (Int.) and rotation (Rot.) are reported for each season-weather condition and aggregated over weather severity levels. Rotation varies only along the Z axis, while X and Y are fixed at their default settings of 0$^\circ$ and $-$46$^\circ$, respectively.\vspace{-0.1cm}}
\begin{tabular}{|l|l|c|c|c|c|}
\hline
 Season & Weather & \multicolumn{4}{c|}{Direct. Light} \\
 & Cond.  & \multicolumn{2}{c|}{Int. (lux)} & \multicolumn{2}{c|}{Rot.~Z (deg)} \\

 &  & min & max & min & max \\\hline
\multirow{3}{*}{spring/summer} & rain & 2.0 & 4.0 & 20 & 300 \\
 & fog & 1.0 & 20.0 & 20 & 189 \\
 & rain$+$fog & 2.0 & 4.0 & 20 & 320 \\\hline
 
\multirow{5}{*}{autumn} & sunny & 1.0 & 3.0 & 83 & 200 \\
 & rain & 1.0 & 2.0 & 20 & 250 \\
 & fog & 1.0 & 2.0 & 200 & 360 \\
 & rain$+$fog & 1.0 & 2.0 & 20 & 250 \\
 & leaves & 1.0 & 3.0 & 83 & 200 \\\hline
 
\multirow{5}{*}{winter} & sunny & 1.0 & 5.0 & 60 & 230 \\
 & rain & 0.8 & 2.0 & 155 & 230 \\
 & fog & 1.0 & 8.0 & 130 & 230 \\
 & rain$+$fog & 1.0 & 2.0 & 150 & 200 \\
 & snow & 0.8 & 2.0 & 155 & 230 \\\hline
\end{tabular}
\end{table}

\begin{table}[t!]
\centering\footnotesize
\caption{Season- and weather-specific Post Process Volume~\cite{Unreal:PostProcessVol} settings. Temp and Tint denote min-max ranges across all environments and weather severity levels.\vspace{-0.1cm}}
\label{tab:SDVW_PostProcessVolume}
\setlength{\tabcolsep}{5pt}
\begin{tabular}{|l|l|r|r|r|r|}
\hline
Season & Weather & \multicolumn{2}{c|}{Temp} & \multicolumn{2}{c|}{Tint} \\
 & Cond. & \multicolumn{1}{c|}{min} & \multicolumn{1}{c|}{max} & \multicolumn{1}{c|}{min} & \multicolumn{1}{c|}{max} \\
\hline
\multirow{4}{*}{spring/summer} 
 & sunny & 4,599 & 5,877 & 0.00 & 0.00 \\
 & rain & 4,300 & 11,240 & $-$0.09 & 0.00 \\
 & fog & 4,300 & 11,439 & 0.00 & 0.00 \\
 & rain$+$fog & 4,300 & 11,240 & $-$0.09 & 0.00 \\\hline

\multirow{5}{*}{autumn} 
 & sunny & 3,900 & 8,768 & 0.00 & 0.00 \\
 & rain & 3,228 & 7,439 & $-$0.02 & 0.00 \\
 & fog & 3,228 & 8,768 & 0.00 & 0.00 \\
 & rain$+$fog & 3,228 & 8,768 & 0.00 & 0.00 \\
 & leaves & 3,900 & 9,181 & 0.00 & 0.00 \\\hline

\multirow{5}{*}{winter} 
 & sunny & 4,884 & 11,493 & 0.00 & 0.05 \\
 & rain & 4,916 & 9,560 & 0.00 & 0.19 \\
 & fog & 3,836 & 11,425 & 0.00 & 0.18 \\
 & rain$+$fog & 5,060 & 9,560 & 0.00 & 0.19 \\
 & snow & 4,160 & 12,804 & $-$0.02 & 0.00 \\\hline
\end{tabular}
\end{table}

The Post Process Volume~\cite{Unreal:PostProcessVol} is applied across all environments to control global color grading. The varied parameters include the temperature type, temperature value (Temp), and green-magenta tint (Tint). The temperature type selects the Temp evaluation mode, either Color Temperature or White Balance. The corresponding Temp and Tint ranges are reported in Tab.~\ref{tab:SDVW_PostProcessVolume}.

\subsection{Statistical Properties}
\begin{figure*}[t!]
\centering
  \includegraphics[width=0.96\textwidth, trim={0.1cm 24.2cm 0.1cm 0cm}, clip]{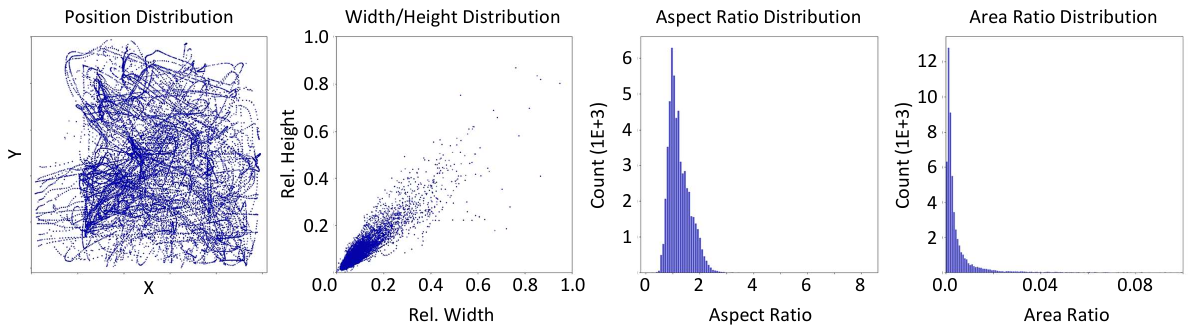}
  \caption{\label{fig:SDVW_Props} Statistical characterization of bounding box annotations in SDV-W. From left to right: spatial distribution of bounding box centers, joint distribution of normalized width and height, distribution of aspect ratios, and distribution of relative bounding box areas. Here, the object area ratio denotes the bounding-box area relative to the image area.}
\end{figure*}

Fig.~\ref{fig:SDVW_Props} visually complements the dataset statistics in Sec.~\ref{subsec:dataset_char} (main paper). Bounding-box centers are distributed broadly across the image plane, while the width-height and area-ratio plots highlight the prevalence of small projected targets. The aspect-ratio distribution reflects the variation in drone pose and viewing geometry preserved in SDV-W.

\section{Object Size Categories}
\begin{table}[t!]
  \caption{\label{tab:obj_size_categories} Definition of object size categories based on bounding-box area $A=w\times h$, extending the COCO taxonomy.\vspace{-0.1cm}}
  \centering\footnotesize
  \begin{tabular}{@{}|l|l|l|@{}}
    \hline
    Size Category & \multicolumn{1}{c|}{Object Area ($\text{pixels}^2$)} & Description\\\hline
    x-small (XS) & \ \ \ \ \ \ \ \  \ \ \ \ \ \ \ $A<16^2$ & extremely small objects\\\hline
    small  (S) & \ \ \ $16^2\leq A<32^2$ & small objects\\\hline
    medium (M) & \ \ \ $32^2\leq A<96^2$ & medium-sized objects\\\hline
    large  (L) & \ \ \ \ \ \ \ \  \ \ \ \ \ \ \ $A\geq96^2$ & large objects\\\hline
  \end{tabular}
\end{table}

For all size-dependent analyses, objects are assigned to four categories according to bounding-box area $A\!\!=\!\!w\times h$ (Tab.~\ref{tab:obj_size_categories}). The definition follows the COCO size taxonomy~\cite{Lin:2014}, with the original S range further subdivided at $A\!=\!16^2$ to introduce an XS category for very small, distant drones. The same categories are used for dataset statistics and size-specific detection results.

\section{Additional Details on Evaluation Datasets}
This section complements Sec.~\ref{subsec:data} (main paper) by presenting the custom seasonal recordings and analyzing the object-scale distributions across both custom recordings and public datasets, including scale shifts induced by resizing to a uniform input resolution of 640$\times$640 pixels.

\subsection{Visual Examples of Custom Recordings}
\begin{figure}[t!]
\centering
  \includegraphics[width=0.47\textwidth, trim={0.1cm 22.8cm 8.4cm 0cm}, clip]{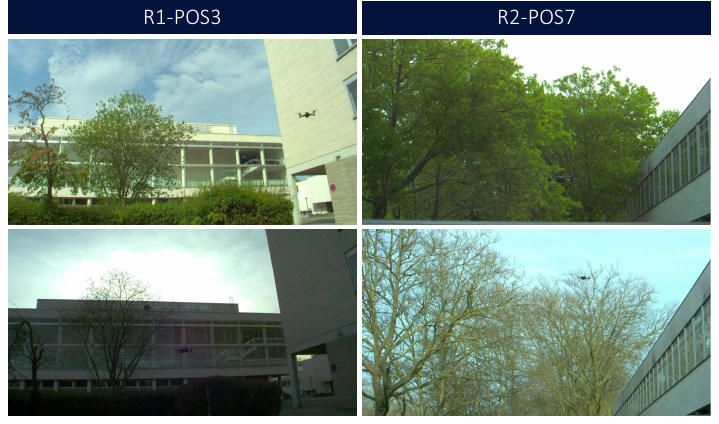}
  \caption{\label{figExamples_custom} Representative frames from the custom recordings R1-POS3 (left) and R2-POS7 (right) under summer (top) and winter (bottom) conditions.}
\end{figure}

Fig.~\ref{figExamples_custom} presents representative summer and winter frames from R1-POS3 and R2-POS7. The fixed camera configurations preserve scene geometry across seasons, with appearance variations primarily reflecting changes in vegetation, from dense summer foliage to largely bare winter trees. Both scenes feature complex, tree-dominated backgrounds that pose challenges for drone detection due to visual clutter and limited target-background contrast.

\begin{table}[t]
\centering\footnotesize
\caption{\label{tab:size_bucket_distribution} Effect of input resizing on object-size category distribution across real-world datasets. Counts are reported before and after resizing to an input resolution of 640$\times$640 pixels. $\Delta$ Share denotes the resizing-induced change in each category's relative frequency expressed in percentage points (pp). ``--'' indicates that no objects of the corresponding size category are present.\vspace{-0.1cm}}
\begin{tabular}{|l|c|c|c|l|}
\hline
Dataset & Size & \multicolumn{2}{c|}{Image Scale} & \multicolumn{1}{c|}{$\Delta$ Share}\\
& Cat. & original & 640$\times$640 & \multicolumn{1}{c|}{(pp)} \\\hline
\multirow{4}{*}{DUT Anti-UAV~\cite{Zhao:2022}} 
& XS & 87  & 569 & $+$21.91\\
& S  & 764 & 674 & $-$4.09\\
& M & 808 & 482 & $-$14.82\\
& L  & 541 & 475 & $-$3.00\\\hline

\multirow{4}{*}{LRDDv1~\cite{Rouhi:2024}} 
& XS & 2,055  & 6,607 & $+$20.37\\
& S  & 6,183  & 5,955 & $-$1.02\\
& M & 10,201 & 8,724 & $-$6.61\\
& L  & 3,903  & 1,056 & $-$12.74\\\hline

\multirow{4}{*}{LRDDv2~\cite{Rouhi:2025}} 
& XS & 4,729  & 9,148 & $+$26.95\\
& S  & 5,916  & 4,746 & $-$7.14\\
& M & 4,493  & 1,651 & $-$17.33\\
& L  & 1,258  & 851   & $-$2.48\\\hline

\multirow{4}{*}{R1-POS3 (S)}
& XS & 118   & 645 & $+$23.43\\
& S  & 626   & 992 & $+$16.27\\
& M & 1,504 & 612 & $-$39.66\\
& L  & 1     & --  & $-$0.04\\\hline

\multirow{4}{*}{R1-POS3 (W)} 
& XS & --    & 335   & $+$13.46\\
& S  & 681   & 1,930 & $+$50.18\\
& M & 1,808 & 224   & $-$63.64\\
& L & -- & --  & \multicolumn{1}{c|}{--} \\ \hline

\multirow{4}{*}{R2-POS7 (S)}
& XS & -- & --  & \multicolumn{1}{c|}{--} \\ 
& S & --   & 20 & $+$1.12\\
& M & 467  & 1,528 & $+$59.41\\
& L & 1,319 & 238  & $-$60.53\\\hline

\multirow{4}{*}{R2-POS7 (W)}
& XS & --    & 2    & $+$0.14\\
& S  & 79    & 1436 & $+$94.11\\
& M & 1,363 & 4    & $-$94.24\\
& L & -- & --  & \multicolumn{1}{c|}{--} \\\hline
\end{tabular}
\end{table}

\begin{figure*}[t!]
\centering
  \includegraphics[width=0.99\textwidth, trim={0.2cm 19.7cm 0.4cm 0cm}, clip]{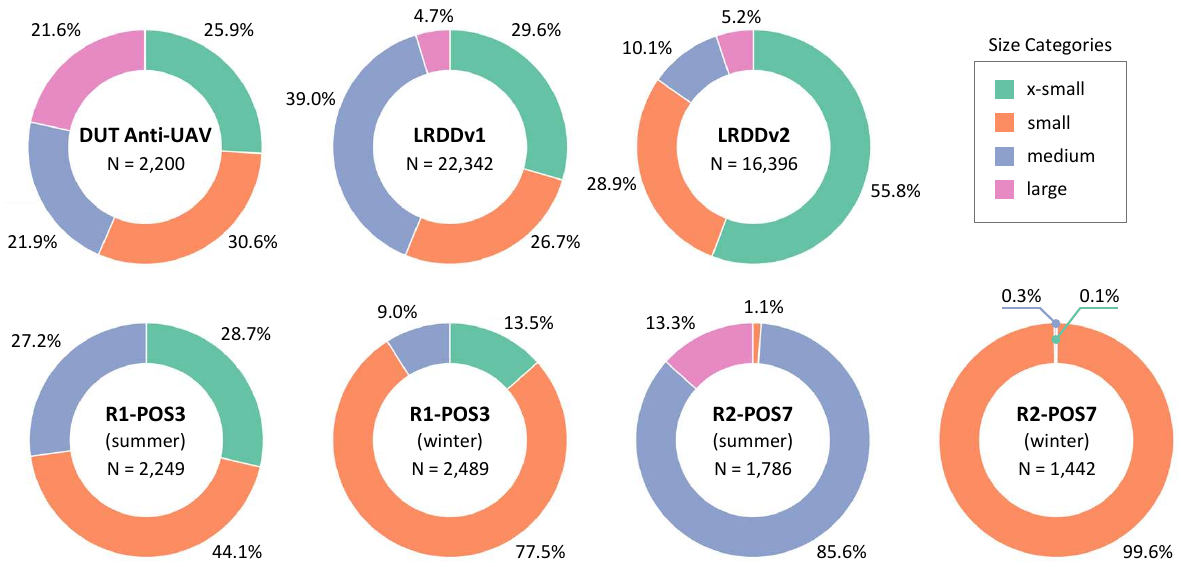}
  \caption{\label{fig:Dist_ObjectSizeCat_Datasets_640x640} Distribution of object size categories across public and custom drone detection datasets. \textbf{Top:} Public datasets DUT Anti-UAV~\cite{Zhao:2022}, LRDDv1~\cite{Rouhi:2024}, and LRDDv2~\cite{Rouhi:2025}. \textbf{Bottom:} Custom datasets R1-POS3 and R2-POS7 under summer and winter conditions. For each dataset, the proportion of x-small, small, medium, and large objects is shown relative to the total number of annotated instances ($N$). Size categories are defined according to Tab.~\ref{tab:obj_size_categories} w.r.t. an input image size of 640$\times$640.}
\end{figure*}

\subsection{Object-Scale Composition}
\label{supp:size_dist} 
Fig.~\ref{fig:Dist_ObjectSizeCat_Datasets_640x640} shows the distribution of annotated instances across the XS/S/M/L categories after mapping all images to the detector's 640$\times$640 input resolution, following the size definitions in Tab.~\ref{tab:obj_size_categories}. The datasets show distinct scale profiles: DUT Anti-UAV~\cite{Zhao:2022} is comparatively balanced across size categories, LRDDv1~\cite{Rouhi:2024} is dominated by medium-size objects, and LRDDv2~\cite{Rouhi:2025} exhibits a pronounced bias toward extra-small instances. The custom recordings R1-POS3 and R2-POS7 are dominated by small-scale objects, with the scale imbalance becoming more pronounced from summer to winter. Large-scale instances are only sparsely represented.

\subsection{Resizing-Induced Scale Compression}
\label{supp:resize_shift}
Resizing native-resolution frames to the detector input compresses object scales in pixel space and systematically transfers instances to smaller size categories. Tab.~\ref{tab:size_bucket_distribution} quantifies this shift through pre- and post-resizing counts and the resulting change in relative frequency ($\Delta$ Share). Across datasets, resizing consistently shifts the scale distribution toward smaller categories, increasing XS/S shares and reducing M/L shares. 

\begin{figure*}[t!]
\centering
  \includegraphics[width=\textwidth, trim={0.1cm 24.5cm 0.1cm 0cm}, clip]{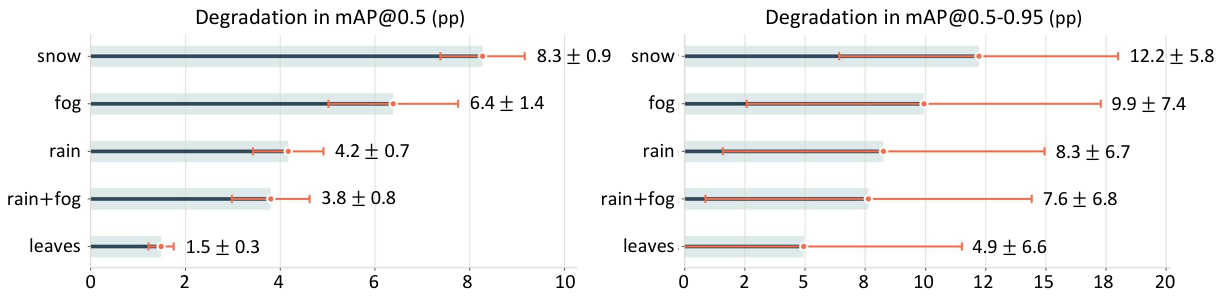}
  \caption{\label{fig:PerfomanceDrop_Weather}Weather-induced degradation of YOLO detection performance relative to clear conditions. Values denote the magnitude of performance loss in percentage points (pp), i.e., the decrease in mAP@0.5 (left) and mAP@0.5-0.95 (right). Larger values indicate stronger degradation for all metrics. Orange markers denote the mean degradation across YOLO model configurations. Horizontal intervals and $\pm$ annotations represent the corresponding (inter-model) standard deviation.}
\end{figure*}

\begin{table*}[t!]
  \caption{\label{tab:results_YOLO-FEDER_PublicData} Impact of weather-conditioned synthetic data (SDV-W) on the performance of YOLO-FEDER FusionNet on LRDDv1~\cite{Rouhi:2024} and DUT Anti-UAV~\cite{Zhao:2022}. Results are reported overall and for each size category defined in Tab.~\ref{tab:obj_size_categories}. Best-performing values are highlighted in \textbf{bold}; tied best values are \underline{underlined}.\vspace{-0.2cm}}
  \centering\footnotesize
  \begin{tabular}{@{}|c|c|c|ccc|c|c|ccc|c|c|@{}}
    \multicolumn{3}{c}{} & \multicolumn{5}{c}{LRDDv1} & \multicolumn{5}{c}{DUT Anti-UAV} \\\hline
    Model & w/ & Size & \multicolumn{3}{c|}{mAP~$\uparrow$} & FNR~$\downarrow$ & FDR~$\downarrow$ & \multicolumn{3}{c|}{mAP~$\uparrow$} & FNR~$\downarrow$ & FDR~$\downarrow$\\
    & SDV-W & Cat. & @0.25 & @0.5 & @0.5-0.95 & & & @0.25 & @0.5 & @0.5-0.95 & &\\\hline
    
    \multirow{9}{*}{YOLO-} & -- & \multirow{2}{*}{all} 
    
    & \textbf{0.772} & \textbf{0.578} & \textbf{0.192} & \textbf{0.310} & 0.056
    & 0.979 & \underline{0.967} & \textbf{0.721} & 0.045 & \underline{0.032}\\
    
    \multirow{9}{*}{FEDER} & \ding{51} & 
    & 0.768 & 0.571 & 0.190 & 0.314 & \textbf{0.054}
    & \textbf{0.981} & \underline{0.967} & 0.716 & \textbf{0.040} & \underline{0.032}\\\cline{2-13}
    
    & -- & \multirow{2}{*}{XS} 
    & \textbf{0.213} & \textbf{0.175} & \textbf{0.060} & \textbf{0.553} & --
    & 0.876 & 0.857 & \textbf{0.500} & 0.086 & --\\
    & \ding{51} &  
    & 0.198 & 0.169 & 0.057 & 0.583 & --
    & \textbf{0.886} & \textbf{0.866} & 0.498 & \textbf{0.077} & --\\\cline{2-13}
    
    &-- & \multirow{2}{*}{S} 
    & 0.631 & \textbf{0.357} & \textbf{0.104} & \underline{0.295} & --
    & 0.933 & \textbf{0.933} & 0.675 & 0.030 & --\\
    & \ding{51} & 
    & \textbf{0.632} & 0.349 & 0.099 & \underline{0.295} & --
    & \textbf{0.935} & 0.931 & \textbf{0.676} & \textbf{0.028} & --\\\cline{2-13}
    
    & -- & \multirow{2}{*}{M} 
    & 0.850 & \underline{0.695} & 0.225 & 0.135 & --
    & 0.949 & 0.948 & \textbf{0.788} & 0.044 & --\\
    & \ding{51} & 
    & \textbf{0.856}  & \underline{0.695} & \textbf{0.226} & \textbf{0.124} & --
    & \textbf{0.956} & \textbf{0.949} & 0.784 & \textbf{0.037} & --\\\cline{2-13}
    
    & -- & \multirow{2}{*}{L} 
    & \textbf{0.595} & 0.586 & \textbf{0.266} & 0.313 &  --
    & \underline{0.972} & 0.960 & \textbf{0.792} & 0.021 & --\\
    & \ding{51} &  
    & 0.593 & \textbf{0.694} & 0.226 & \textbf{0.124} & --
    & \underline{0.972} & \textbf{0.962} & 0.790 & \textbf{0.015} & --\\\hline\noalign{\smallskip}
    \multicolumn{3}{l}{\ding{51} = applies}\vspace{-0.2cm}
  \end{tabular}
\end{table*}

\section{Extended Quantitative Results}
This section provides a more detailed breakdown of the quantitative results summarized in Sec.~\ref{sec:results} (main paper).

\subsection{Weather Degradation under Stricter IoU Thresholds}
Fig.~\ref{fig:PerfomanceDrop_Weather} complements the analysis in Sec.~\ref{subsec:degradation_SDV-W} (main paper) by evaluating degradation under stricter IoU thresholds, using mAP@0.5 and mAP@0.5-0.95. Degradation is reported as magnitude of performance loss in percentage points (pp) as mean $\pm$ standard deviation across YOLO configurations. The condition ranking remains consistent across thresholds, with snow causing the largest drop, followed by fog, rain, rain$+$fog, and leaves. At mAP@0.5, degradation remains close to the mAP@0.25 results, whereas mAP@0.5-0.95 reveals a stronger localization-sensitive penalty across all conditions. Variability across YOLO configurations increases under mAP@0.5-0.95, with standard deviations rising to 5.8-7.4\,pp compared with $\le$1.4\,pp at mAP@0.5. Thus, the degradation magnitude seems to be architecture-dependent, while the condition ordering remains stable across models.

\subsection{Detailed Analysis of SDV-W-Based Training}
This section provides a more detailed assessment of the effects of SDV-W-based training (cf.~Sec.~\ref{subsec:res_training_impact}, main paper). 

\vspace{-0.3cm}
\paragraph{Size-Resolved Behavior on LRDDv1 \& DUT Anti-UAV.} For YOLO-FEDER FusionNet, incorporating SDV-W into 
the training composition leaves aggregate LRDDv1~\cite{Rouhi:2024} performance essentially unchanged across mAP criteria, yet resolves into modest size-dependent variations (see Tab.~\ref{tab:results_YOLO-FEDER_PublicData}). The most pronounced scale-specific gains occur for large objects, with mAP@0.5 increasing by 10.8\,pp and FNR decreasing by 18.9\,pp. Medium objects exhibit only a marginal FNR reduction (FNR: $-$1.1\,pp), small objects remain effectively unchanged, and extra-small objects degrade slightly (mAP@0.5: $-$0.6\,pp; FNR: $+$3.0\,pp), partially offsetting the large-object gains. On DUT Anti-UAV~\cite{Zhao:2022}, SDV-W yields a marginally favorable trend: mAP remains essentially stable across size categories ($|\Delta|\leq$~0.7\,pp), while FNR decreases slightly but consistently (XS: $-$0.9, S: $-$0.2, M: $-$0.7, L: $-$0.6\,pp; aggregate: $-$0.5\,pp). Thus, the effect is confined to marginal detection gains rather than improved localization.

\vspace{-0.3cm}
\paragraph{Architecture-Invariant Effects on LRDDv2.}
Across all standard YOLO variants, the impact of SDV-W on mAP is architecture-dependent (see Tab.~\ref{tab:results_YOLOs_LRDD_v2}): four variants improve (v5l, v8l, v9c, v11l; up to $+$3.2\,pp at mAP@0.5), while three degrade (v8m, v9e, v11x), yielding the near-neutral mean reported in the main paper. The architecture-consistent pattern is instead a reduction in false detections, with FDR decreasing across all variants, most prominently for v8l (0.098$\rightarrow$0.048) and v9c (0.174$\rightarrow$0.121). Since FNR remains high ($\geq$0.61) and largely unaffected, mAP is of limited interpretive value on this dataset. The consistent FDR reduction constitutes the principal architecture-independent benefit of SDV-W.

\begin{table*}[t!]
  \caption{\label{tab:results_YOLOs_LRDD_v2}Comparison of training with and without weather-condition synthetic data (SDV-W) on LRDDv2~\cite{Rouhi:2025} across multiple YOLO architectures and model scales. Best-performing values are highlighted in \textbf{bold}, while tied best values are \underline{underlined}.\vspace{-0.2cm}}
  \centering\footnotesize
  \begin{tabular}{@{}|l|c|ccc|c|c|@{}}
    \hline
     \multicolumn{1}{|c|}{YOLO} & w/ & \multicolumn{3}{c|}{mAP~$\uparrow$} & FNR~$\downarrow$ & FDR~$\downarrow$ \\
    & SDV-W & @0.25 & @0.5 & @0.5-0.95 & & \\\hline
    \multirow{2}{*}{\ \ v5l} 
    & -- &  0.517 & 0.495 & 0.255 & 0.659 & 0.091\\
    & \ding{51} & \textbf{0.530} & \textbf{0.509} & \textbf{0.266} & \textbf{0.653} & \textbf{0.083}\\\hline
    
    \multirow{2}{*}{\ \ v8m} 
    & -- & \textbf{0.534} & \textbf{0.514} & \underline{0.266} & \textbf{0.630} & 0.108\\
    & \ding{51} & 0.519 & 0.504 & \underline{0.266} & 0.642 & \textbf{0.084}\\\hline
    
    \multirow{2}{*}{\ \ v8l} 
    & -- & 0.516 & 0.498 & 0.252 & 0.646 & 0.098\\	
    & \ding{51} & \textbf{0.524} & \textbf{0.504} & \textbf{0.254} & \textbf{0.645} & \textbf{0.048}\\\hline
    
    \multirow{2}{*}{\ \ v9c} 
    & -- & 0.477 & 0.464 & 0.232 & 0.657 & 0.174\\	
    & \ding{51} & \textbf{0.515} & \textbf{0.496} & \textbf{0.248} & \textbf{0.615} & \textbf{0.121}\\\hline
    
    \multirow{2}{*}{\ \ v9e} 
    & -- & \textbf{0.515} & \textbf{0.495} & \textbf{0.257} & 0.643 & 0.047\\	
    & \ding{51} & 0.498 & 0.481 & 0.253 & \textbf{0.641} & \textbf{0.044}\\\hline	
    
    \multirow{2}{*}{\ \ v11l} 
    & -- & 0.533 & 0.514 & 0.268 & \underline{0.614} & 0.057\\	
    & \ding{51} & \textbf{0.551} & \textbf{0.534} & \textbf{0.279} & \underline{0.614} & \textbf{0.044}\\\hline	
    
    \multirow{2}{*}{\ \ v11x} 
    & -- & \textbf{0.546} & \textbf{0.525} & \textbf{0.276} & \textbf{0.621} & 0.094\\	
    & \ding{51} & 0.527	& 0.504 & 0.255 & 0.626 & \textbf{0.067}\\\hline\noalign{\smallskip}
    \multicolumn{3}{l}{\ding{51} = applies} \vspace{-0.2cm}
  \end{tabular}
\end{table*}

\begin{table*}[t!]
  \caption{\label{tab:results_YOLO-FEDER_R1_R2_2}Performance of YOLO-FEDER FusionNet with and without weather-conditioned synthetic training data (SDV-W) under seasonal conditions (summer~vs.~winter) on the datasets R1-POS3 and R2-POS7. Results are reported overall and for each size category defined in Tab.~\ref{tab:obj_size_categories}. Best-performing values are highlighted in \textbf{bold}; tied best values are \underline{underlined}. ``--'' indicates that no objects of the corresponding size category are present.\vspace{-0.4cm}}
  \centering\footnotesize
  \begin{tabular}{@{}|l|c|c|ccc|c|c|ccc|c|c|@{}}
     \multicolumn{3}{c}{} & \multicolumn{5}{c}{Summer} & \multicolumn{5}{c}{Winter}\\\hline
    Dataset & w/ & \multicolumn{1}{c|}{Size} & \multicolumn{3}{c|}{mAP~$\uparrow$} & FNR~$\downarrow$ & FDR~$\downarrow$ & \multicolumn{3}{c|}{mAP~$\uparrow$} & FNR~$\downarrow$ & FDR~$\downarrow$\\
    & SDV-W & Cat. & @0.25 & @0.5 & @0.5-0.95 & & & @0.25 & @0.5 & @0.5-0.95 & &\\\hline
        
    \multirow{8}{*}{R1-POS3}
    & -- & \multirow{2}{*}{all} 
    & 0.764 & 0.743 & 0.331 & \textbf{0.313} & 0.083
    & 0.931 & 0.827 & \textbf{0.404} & 0.208 & 0.006\\
    & \ding{51} & 
    & \textbf{0.821} & \textbf{0.794} & \textbf{0.354} & 0.326 & \textbf{0.004} 
    & \textbf{0.965} & \textbf{0.836} & 0.393 & \textbf{0.164} & \textbf{0.004}\\\cline{2-13}
    
    & -- & \multirow{2}{*}{XS} 
    & 0.364 & \textbf{0.358} & \textbf{0.175} & \textbf{0.555} & --
    & 0.596 & 0.572 & 0.250 & 0.206 & --\\
    & \ding{51} & 
    & \textbf{0.373} & 0.348 & 0.160 & 0.575 & --
    & \textbf{0.644} & \textbf{0.634} & \textbf{0.271} & \textbf{0.191} & --\\\cline{2-13}
    
    & -- & \multirow{2}{*}{S} 
    & 0.727 & 0.713 & 0.345 & \textbf{0.268} & --
    & 0.890 & 0.764 & \textbf{0.375} & 0.224 & --\\
    & \ding{51} & 
    & \textbf{0.792} & \textbf{0.772} & \textbf{0.367} & 0.275 & --
    & \textbf{0.938} & \textbf{0.768} & 0.361 & \textbf{0.170} & --\\\cline{2-13}
    
    & -- & \multirow{2}{*}{M} 
    & 0.860 & 0.848 & 0.358 & \textbf{0.132} & -- 
    & 0.898 & 0.828 & 0.467 & 0.076 & -- \\
    & \ding{51} & 
    & \textbf{0.897} & \textbf{0.889} & \textbf{0.377} & 0.147 & --
    & \textbf{0.926} & \textbf{0.875} & \textbf{0.479} & \textbf{0.063} & --\\\hline
    
    \multirow{10}{*}{R2-POS7}
    & -- & \multirow{2}{*}{all}
    & \textbf{0.985} & \textbf{0.909} & \textbf{0.450} & 0.076 & 0.010
    & 0.983 & \textbf{0.749} & \textbf{0.294} &  \textbf{0.025} & 0.054\\
    & \ding{51} & 
    & 0.982 & 0.901 & 0.422 & \textbf{0.061} & \textbf{0.008}
    & \textbf{0.988} & 0.675 & 0.223 & 0.028 & \textbf{0.006}\\\cline{2-13}
    
    & -- & \multirow{2}{*}{XS} 
    & -- & -- & -- & -- & -- 
    & \textbf{0.249} & \underline{0.000} & \underline{0.000} & \underline{0.000} & --\\
    & \ding{51} & 
    & -- & -- & -- & -- & -- 
    & 0.077 & \underline{0.000} & \underline{0.000} & \underline{0.000} & --\\\cline{2-13}
    
    & -- & \multirow{2}{*}{S} 
    & \textbf{0.047} & 0.022 & \textbf{0.009} & \underline{0.700} & --
    & 0.982 & \textbf{0.748} & \textbf{0.293} & \textbf{0.025} & --\\
    & \ding{51} & & 0.030 & \textbf{0.030} & 0.008 & \underline{0.700} & --
    & \textbf{0.985} & 0.674 & 0.222 & 0.028 & --\\\cline{2-13}
    
    & -- & \multirow{2}{*}{M} 
    & \textbf{0.979} & \textbf{0.887} & \textbf{0.419} & 0.079 & -- 
    & 0.463 & 0.463 & 0.336 & \underline{0.000} & --\\
    & \ding{51} & & 0.975 & 0.878 & 0.394 & \textbf{0.062} & --
    & \textbf{0.794} & \textbf{0.794} & \textbf{0.506} & \underline{0.000} & --\\\cline{2-13}
    
    & -- & \multirow{2}{*}{L} 
    & \underline{0.991} & \underline{0.991} & \textbf{0.570} & 0.008 & -- 
    & -- & -- & -- & -- & --\\
    & \ding{51} & 
    & \underline{0.991} & \underline{0.991} & 0.533 & \textbf{0.000} & --
    & -- & -- & -- & -- & --\\\hline\noalign{\smallskip}
    \multicolumn{8}{l}{\ding{51} = applies} \vspace{-0.2cm}

  \end{tabular}
\end{table*}

\vspace{-0.3cm}
\paragraph{Size- \& Season-Resolved Behavior on Custom Recordings.}
On R1-POS3, the improvements observed under summer conditions are scale-selective, with mAP@0.5 gains concentrated in small and medium object categories ($+$5.9 and $+$4.1\,pp), while performance for extra-small objects shows a marginal negative trend (see Tab.~\ref{tab:results_YOLO-FEDER_R1_R2_2}). Under winter conditions, SDV-W-based training yields consistent improvements in mAP@0.25 and mAP@0.5 across all size categories. Performance at stricter IoU thresholds is slightly lower but broadly comparable to training without SDV-W. FNR is also modestly reduced, indicating a small but consistent reduction in missed detections. On dataset R2-POS7, the size-resolved results provide a more nuanced interpretation of the mAP decline under winter conditions: while the localization of medium-size objects improves significantly (mAP@0.5: 0.463$\rightarrow$0.794; mAP@0.5-0.95: $+$17.0\,pp), mAP decreases for small and extra-small objects. Since small objects dominate the size distribution, their degradation outweighs the medium-scale gains and drives the aggregate mAP decline. FDR decreases consistently across both seasons.

\begin{table*}[t!]
  \caption{\label{tab:ablation_study}Ablation study disentangling the effects of weather diversity and training set size for YOLO-FEDER FusionNet. Weather-conditioned synthetic data (SDV-W) either substitute a subset of SDV samples or extend the original training set, consisting of SDV and DUT Anti-UAV~\cite{Zhao:2022}.\vspace{-0.2cm}}
  \centering\footnotesize
  \begin{tabular}{@{}|l|c|ccc|c|c|@{}}
    \hline
    Dataset & w/ & \multicolumn{3}{c|}{mAP~$\uparrow$} & \multicolumn{1}{c|}{FNR~$\downarrow$} & \multicolumn{1}{c|}{FDR~$\downarrow$}\\
    & SDV-W & \multicolumn{1}{c}{@0.25} & \multicolumn{1}{c}{@0.5} & \multicolumn{1}{c|}{@0.5-0.95} & &\\\hline
    
    \multirow{3}{*}{LRDDv1~\cite{Rouhi:2024}} 
    & -- & 0.772 & 0.578 & 0.192 & 0.310 & 0.056\\
    & \ding{51}$^R$ & $0.774 \pm 0.008$ & $0.577 \pm 0.010$ & $0.192 \pm 0.005$ & $0.315 \pm 0.006$ & $0.053 \pm 0.011$\\
    & \ding{51}$^A$ & 0.768 & 0.571 & 0.190 & 0.314 & 0.054 \\\hline

    \multirow{3}{*}{LRDDv2~\cite{Rouhi:2025}} 
    & -- & 0.540 & 0.507 & 0.254 & 0.515 & 0.270\\
    & \ding{51}$^R$ & $0.551 \pm 0.002$ & $0.516 \pm 0.004$ & $0.253 \pm 0.002$ & $0.510 \pm 0.006$ & $0.239 \pm 0.022$\\
    & \ding{51}$^A$ & 0.561 & 0.530 & 0.258 & 0.491 & 0.235\\\hline
    
    \multirow{3}{*}{DUT Anti-UAV~\cite{Zhao:2022}} 
    & -- & 0.979 & 0.967 & 0.721 & 0.045 & 0.032\\
    & \ding{51}$^R$ & $0.978 \pm 0.001$ & $0.962 \pm 0.002$ & $0.715 \pm 0.003$ & $0.044 \pm 0.005$ & $0.032 \pm 0.003$\\
    & \ding{51}$^A$ & 0.981 & 0.967 & 0.716 & 0.040 & 0.032\\\hline
    
    \multirow{3}{*}{R1-POS3 (S)} 
    & -- & 0.764 & 0.743 & 0.331 & 0.313 & 0.083\\
    & \ding{51}$^R$ & $0.813 \pm 0.020$ & $0.787 \pm 0.021$ & $0.363 \pm 0.014$ & $0.338 \pm 0.018$ & $0.006 \pm 0.002$\\
    & \ding{51}$^A$ & 0.821 & 0.794 & 0.354 & 0.326 & 0.004\\\hline

    \multirow{3}{*}{R1-POS3 (W)} 
    & -- & 0.931 & 0.827 & 0.404 & 0.208 & 0.006\\
    & \ding{51}$^R$ & $0.938 \pm 0.007$ & $0.842 \pm 0.020$ & $0.369 \pm 0.029$ & $0.192 \pm 0.025$ & $0.005 \pm 0.002$\\
    & \ding{51}$^A$& 0.965 & 0.836 & 0.393 & 0.164 & 0.004\\\hline

    \multirow{3}{*}{R2-POS7 (S)} 
    & -- & 0.985 & 0.909 & 0.450 & 0.076 & 0.010\\
    & \ding{51}$^R$ & $0.979 \pm 0.004$ & $0.897 \pm 0.009$ & $0.421 \pm 0.007$ & $0.092 \pm 0.017$ & $0.011 \pm 0.005$\\
    & \ding{51}$^A$ & 0.982 & 0.901 & 0.422 & 0.061 & 0.008\\\hline
    
    \multirow{3}{*}{R2-POS7 (W)} 
    & -- & 0.983 & 0.749 & 0.294 & 0.025 & 0.054 \\
    & \ding{51}$^R$ & $0.988 \pm 0.006$ & $0.736 \pm 0.074$ & $0.288 \pm 0.050$ & $0.025 \pm 0.009$ & $0.010 \pm 0.009$\\
    & \ding{51}$^A$ & 0.988 & 0.675 & 0.223 & 0.028 & 0.006\\\hline\noalign{\smallskip}
    \multicolumn{7}{l}{\ding{51}$^R$\hspace{0.3cm} Replacement of randomly selected SDV samples with SDV-W, keeping the training set size constant (mean $\pm$ std over 5 runs).}\\
    \multicolumn{7}{l}{\ding{51}$^A$\hspace{0.3cm} Extension of the original training set with SDV-W.} \vspace{-0.2cm}
  \end{tabular}
\end{table*}

\subsection{Ablation Details for Weather Diversity vs. Training-Set Size}
\label{subsec:ablation}
Tab.~\ref{tab:ablation_study} compares three training configurations that differ in the use of SDV-W: a baseline without SDV-W (--), additive inclusion of SDV-W ($\checkmark^A$), and size-preserving replacement with \mbox{SDV-W} ($\checkmark^R$). In the replacement setting, \mbox{SDV-W} replaces an equally sized, randomly sampled subset of the original SDV images from the baseline training composition, preserving the baseline training-set size. Repeating this procedure over five independently sampled subsets yields five YOLO-FEDER FusionNet models, with results reported as mean $\pm$ standard deviation.

Across the public datasets, the replacement setting remains largely comparable to the baseline, while additive inclusion yields the more consistently competitive configuration. This suggests that SDV-W is most effective when added as complementary coverage rather than substituted for existing SDV samples. On the custom recordings, however, replacement captures much of the additive-setting benefit despite the fixed training-set size, indicating that the gains are primarily attributable to improved domain alignment rather than increased data volume alone.

\begin{table}[t!]
  \caption{\label{tab:results_tiling_LRDDv2} Impact of tiled inference relative to resize-based inference for YOLO-FEDER FusionNet on LRDDv2~\cite{Rouhi:2025}. Best-performing values are highlighted in \textbf{bold}.\vspace{-0.2cm}}
  \centering\footnotesize
  \begin{tabular}{@{}|c|c|ccc|c|c|@{}}
    \hline
    Tiling & Size & \multicolumn{3}{c|}{mAP~$\uparrow$} & FNR~$\downarrow$ & FDR~$\downarrow$\\
     & Cat. & @0.25 & @0.5 & @0.5-0.95 & &\\\hline

    -- & \multirow{2}{*}{all} & 0.561 & 0.530 & \textbf{0.258} & 0.491 & \textbf{0.235}\\
    \ding{51} & & \textbf{0.669} & \textbf{0.580} & 0.245 & \textbf{0.305} & 0.426\\\hline

    -- & \multirow{2}{*}{XS} & 0.152 & 0.133 & 0.055 & 0.735 & --\\
    \ding{51} & & \textbf{0.407} & \textbf{0.312} & \textbf{0.108} & \textbf{0.434} & --\\\hline

    -- & \multirow{2}{*}{S} & 0.705 & 0.690 & 0.311 & 0.227 & --\\
    \ding{51} & & \textbf{0.756} & \textbf{0.708} & \textbf{0.321} & \textbf{0.172} & --\\\hline

    -- & \multirow{2}{*}{M} & 0.762 & \textbf{0.755} & \textbf{0.448} & 0.138 & --\\
    \ding{51} & & \textbf{0.766} & 0.747 & 0.387 & \textbf{0.098} & --\\\hline

    -- & \multirow{2}{*}{L} & \textbf{0.856} & \textbf{0.855} & \textbf{0.600} & \textbf{0.031} & --\\
    \ding{51} & & 0.708 & 0.661 & 0.287 & 0.062 & --\\\hline\noalign{\smallskip}
    \multicolumn{3}{l}{\ding{51} = tiled inference} \vspace{-0.2cm}
  \end{tabular}
\end{table}

\subsection{Tiled Inference on LRDDv2}
\label{subsec:tiling_supp}
This section provides further details on the tiled-inference evaluation presented in Sec.~\ref{subsec:res_tiling} (main paper). The analysis of detection performance across object scales (Sec.~\ref{subsec:res_training_impact}, main paper) identifies extra-small drones as the primary source of detection failures, with SDV-W-based training yielding only marginal improvements (XS mAP@0.5 on LRDDv2: 0.125$\rightarrow$0.133). This suggests that the primary performance bottleneck stems from limited spatial resolution rather than insufficient training data. Under standard YOLO-based inference, high-resolution frames are globally resized to 640$\times$640 pixels and progressively downsampled into spatially coarse, semantically rich feature maps, leaving limited spatial detail for tiny-object detection. In contrast, tiled inference offers an established scale-preserving alternative by partitioning frames into overlapping tiles extracted at native resolution (\eg, 640$\times$640), processing them independently, and fusing the resulting detections in the original image space after reprojection. 

To assess the role of spatial resolution, tiled inference is coupled with YOLO-FEDER FusionNet (trained on the combination of SDV~\cite{Lenhard:2025}, DUT Anti-UAV~\cite{Zhao:2022}, and SDV-W). Evaluation is conducted on LRDDv2~\cite{Rouhi:2025}, which exhibits the strongest concentration of small and extra-small objects (see~Fig.~\ref{fig:Dist_ObjectSizeCat_Datasets_640x640}). If spatial resolution is the primary bottleneck, preserving native object scales should yield greater XS detection gains than SDV-W-based training alone. 

This experiment provides an initial assessment of the applicability of tiled inference to extra-small drone detection in the broader context of urban surveillance. A systematic exploration of the tiling design space is beyond the scope of this work.

\vspace{-0.3cm}
\paragraph{Parameter Selection.} Following SAHI~\cite{Akyon:2022}, the tiling configuration is defined by the tile size, inter-tile overlap, and the post-processing strategy used to merge detections across tiles. The tile size is set to the YOLO default input resolution of 640$\times$640 pixels. Duplicate predictions across tiles are resolved with SAHI's default post-processing configuration: greedy non-maximum merging with intersection over smaller area (IoS) matching at a threshold of 0.5. Under the square, padding-free input constraint of YOLO-FEDER FusionNet (see~\cite{Lenhard:2024,Lenhard:2026} for details), the effective inter-tile overlap is determined by the original image dimensions relative to the fixed tile size. Therefore, a minimum adjacent-tile overlap of 0.2 (SAHI default) is imposed, with per-image adjustment to ensure complete spatial coverage.

\vspace{-0.3cm}
\paragraph{Results on LRDDv2.} 
Tiled inference reveals a pronounced scale-dependent trade-off (cf.~Tab.~\ref{tab:results_tiling_LRDDv2}). For extra-small drones, mAP@0.25 increases from 0.152 to 0.407 and mAP@0.5 from 0.133 to 0.312, accompanied by 30.1\,pp reduction in FNR. Small drones similarly exhibit consistent performance gains, with improved localization accuracy and reduced FNR. For medium-sized drones, FNR decreases, but mAP declines, particularly under stricter IoU criteria. In contrast, tiled inference substantially degrades large-object detection performance, as drones extending beyond tile boundaries are prone to fragmentation: mAP@0.5 drops from 0.855 to 0.661, mAP@0.5-0.95 from 0.600 to 0.287, and FNR doubles. 

The improvement in XS mAP@0.5 ($+$17.9\,pp) exceeds that achieved through SDV-W-based training ($+$0.8\,pp) by more than an order of magnitude, suggesting that residual detection failures at this scale are predominantly resolution-limited. Nevertheless, the pronounced degradation in large-object detection, elevated image-level FDR (0.235$\rightarrow$0.426), and increased computational overhead constrain the viability of uniform tiling as a direct replacement for global inference in real-time urban drone surveillance.

\section{Scope, Limitations, and Future Directions}
\label{sec:scope_and_limitations}
This section expands on the discussion in Sec.~\ref{sec:conclusion} of the main paper, highlighting the scope and limitations of SDV-W and the experimental evaluation, alongside potential directions for future research.

\vspace{-0.3cm}
\paragraph{Environmental \& Seasonal Coverage.}
SDV-W encompasses three urban environments, inheriting geometric diversity, object distributions, and camera viewpoints from SDV. Weather and seasonal variations are introduced under fixed scene geometry, enabling controlled comparisons across environmental conditions. The focus on urban drone surveillance emphasizes settings in which vegetation provides prominent seasonal cues. Extending SDV-W to additional urban and non-urban environments would further enhance scene diversity and the representation of weather and seasonal appearance variations across different environmental contexts, and is therefore a direction for future work.

\vspace{-0.3cm}
\paragraph{Weather Rendering Scope.}
SDV-W employs scene-level weather rendering to enable controlled and reproducible variations in environmental conditions. Camera-specific effects, including lens droplets, condensation, and sensor noise, are not explicitly modeled in SDV-W. Incorporating such effects in future extensions could further enhance visual diversity and fidelity to real-world imaging conditions.

\vspace{-0.3cm}
\paragraph{Real-World Validation.}
Real-world evaluation is constrained by the limited availability of drone datasets featuring visually observable adverse weather. Reported weather conditions often characterize general environmental or surface states, such as snow-covered ground, rather than directly observable atmospheric effects (Tab.~1, main paper). Consequently, real-world transfer is primarily evaluated under clear and overcast conditions, while controlled synthetic experiments enable a complementary assessment of weather-specific performance degradation. Extending the evaluation to real-world adverse weather would provide further insights into the transferability of SDV-W's observed benefits to genuine atmospheric conditions. Accordingly, a necessary next step is to collect real-world drone recordings under adverse weather conditions to directly assess whether the observed benefits of SDV-W transfer to real-world atmospheric conditions.

\vspace{-0.3cm}
\paragraph{Detector Architecture.} The evaluation focuses on CNN-based detectors from the YOLO family, providing a well-established and practically relevant foundation for systematically investigating weather-induced performance degradation in drone-based object detection.  Transformer-based architectures, such as RT-DETR~\cite{Zhao:2024,Lv:2024}, employ different feature extraction and aggregation mechanisms that may influence their sensitivity to spatially extended degradations, including fog and heavy precipitation. Extending the evaluation to these architectures would enable a systematic investigation of architecture-dependent weather sensitivity and the broader applicability of SDV-W-based training.

\vspace{-0.3cm}
\paragraph{Tiled Inference.}
The tiled-inference evaluation uses SAHI's default configuration, providing an initial assessment of its impact on detection performance. Alternative tile sizes, overlap ratios, and prediction-merging strategies may influence the trade-off between extra-small and larger objects. Future investigations will explore scale-adaptive approaches, including global-tiled prediction fusion and selective tiling, to retain small-object detection gains while reducing performance trade-offs and computational overhead.

\end{document}